\documentclass{article}
\usepackage{graphicx}
\usepackage{bm}
\usepackage{xcolor}
\usepackage[margin=1in]{geometry}
\usepackage{amsmath}
\usepackage{amssymb}
\usepackage{setspace}
\usepackage{caption} 
\usepackage[natbibapa]{apacite}
\usepackage{booktabs}
\usepackage{graphicx}
\usepackage{subcaption}
\usepackage{svg}
\svgpath{{figures/}}

\usepackage{natbib}
\usepackage{tabularx}
\usepackage{array}
\usepackage{threeparttable}
\usepackage{multirow}
\usepackage{xr}
\usepackage{placeins}
\usepackage{caption} 
\usepackage{amsmath,calc}
\usepackage{subcaption}
\usepackage{comment}
\usepackage[title]{appendix}
\title{Learning Preference from Observed Rankings}
\author{
Yu-Chang Chen \\
National Taiwan University
\and
Chen Chian Fuh \\
National Taiwan University
\and
Shang En Tsai \\
National Taiwan University
}
\date{This Version: January 2026}

\begin{document}

\maketitle
\begin{abstract}
\noindent Estimating consumer preferences is central to many problems in economics and marketing. This paper develops a flexible framework for learning individual preferences from partial ranking information by interpreting observed rankings as collections of pairwise comparisons with logistic choice probabilities. We model latent utility as the sum of interpretable product attributes, item fixed effects, and a low-rank user-item factor structure, enabling both interpretability and information sharing across consumers and items. We further correct for selection in which comparisons are observed: a comparison is recorded only if both items enter the consumer’s consideration set, inducing exposure bias toward frequently encountered items. We model pair observability as the product of item-level observability propensities and estimate these propensities with a logistic model for the marginal probability that an item is observable. Preference parameters are then estimated by maximizing an inverse-probability-weighted (IPW), ridge-regularized log-likelihood that reweights observed comparisons toward a target comparison population. To scale computation, we propose a stochastic gradient descent (SGD) algorithm based on inverse-probability resampling, which draws comparisons in proportion to their IPW weights. In an application to transaction data from an online wine retailer, the method improves out-of-sample recommendation performance relative to a popularity-based benchmark, with particularly strong gains in predicting purchases of previously unconsumed products.
\end{abstract}
\noindent\textbf{Keywords:} incomplete rank data; exploded logit; inverse probability weighting; ridge regularization.
\par

\section{Introduction}
\noindent Estimating consumer preferences is a foundational task in economics. Preference estimates are central inputs to structural demand and discrete-choice models that quantify substitution patterns and willingness-to-pay, enabling counterfactual evaluation of pricing, assortment, and new-product decisions \citep{mcfadden1973conditional, train2009discrete}. They also underpin stated-preference methods such as conjoint and choice experiments, which remain workhorse tools for product design, positioning, and market simulation when historical sales data are limited or unavailable \citep{green1990conjoint, louviere2000stated}. In digital marketplaces, learning preferences at the individual level is equally critical for personalization: recommender systems, search ranking, and targeted promotions , often leveraging latent-factor representations and pairwise ranking objectives \citep{koren2009matrix,koren2021advances}. 

This paper develops a new approach to learning individual preferences from ranking data. We consider settings in which researchers observe how consumers rank subsets of available options, rather than complete orderings over the entire choice set. Such ranking data may be obtained directly from surveys that ask respondents to compare or rank alternatives, or indirectly inferred from revealed preference in observed choices. For example, if a consumer chooses item \(j\) when item \(j'\) is also available, this choice reveals that \(j\) is preferred to \(j'\). Crucially, we do not assume that researchers observe a complete ranking over all items for any individual. Instead, the central objective of this paper is to recover consumers' underlying preference structures—and, in particular, to infer rankings over unobserved item pairs—using only partial and incomplete ranking information observed across individuals.

The proposed approach learns individual preferences by exploiting two complementary sources of information. First, preferences can be inferred from how consumers rank items with different observable attributes. Rankings among observed items reveal consumers' tastes over product characteristics, which can then be extrapolated to items that have not been directly ranked. For example, if a consumer systematically ranks Brand~A above Brand~B among the items she evaluates, this pattern suggests that products sharing Brand~A’s attributes are likely to be preferred. 

Second, the approach leverages similarities in observed rankings across consumers. When two consumers exhibit similar ranking patterns over a subset of items, information about one consumer's preferences can inform about the other's unobserved comparisons. In particular, if two consumers share closely aligned rankings on a common set of items and one consumer is observed to prefer item~A to item~B, it is more likely that the other consumer also prefers item~A to item~B. By combining attribute-based extrapolation with cross-consumer similarity, the method pools information efficiently to recover preferences beyond the directly observed rankings.

To operationalize these ideas and estimate preferences, we model consumers' latent utility and interpret the observed ranking data as a collection of pairwise choice comparisons. The utility of an item is decomposed into an interpretable component that captures systematic preferences over observable product attributes—such as brand or country of origin—and a latent factor component that captures residual similarities between users and items not explained by these attributes. This structure allows the model to combine both attribute-based extrapolation with information pooled across consumers who exhibit similar ranking patterns.
Estimation proceeds by viewing each observed ranking as the outcome of a hypothetical binary choice. When item \(A\) is observed to be ranked above item \(B\) for a given consumer, we interpret this as evidence that the consumer would choose \(A\) over \(B\) if both items were simultaneously available. To map this interpretation into an estimable likelihood, we impose the standard assumption that the idiosyncratic utility term follow a type-I extreme value distribution. Under this assumption, the probability that item \(A\) is preferred to item \(B\) takes a logistic form, and preference estimation reduces to a binary response problem with a logistic specification. Because the number of implied pairwise comparisons can be very large, we estimate the model using a stochastic optimization procedure: in each iteration, we randomly sample a subset of observed ranking pairs and update the parameters using gradient-based steps in the spirit of stochastic gradient descent. This approach enables scalable estimation while efficiently exploiting the information contained in large and sparse ranking datasets.

We illustrate the proposed approach using transaction data from an online wine retailer. We begin by aggregating individual wine products into interpretable categories defined by region of origin (e.g., Bordeaux), grape variety (e.g., Cabernet Sauvignon), and price range (e.g., \$15–30 USD). Based on these categories, we construct customer-specific ranking data from observed purchase behavior. Specifically, if a customer has purchased a particular wine category but has not purchased another, we interpret this pattern as revealing a preference for the purchased category over the unpurchased one. 
We then apply our method to these constructed rankings to recover individual-level preference structures. We assess the practical usefulness of the approach by evaluating its out-of-sample predictive performance in recommendation tasks. Beyond prediction, the empirical application also demonstrates how the estimated preference rankings can be translated into actionable managerial insights. In particular, by computing composition and percentile lifts from model-implied preferences, we show how the framework can inform targeting and segmentation decisions in marketing practice.

\subsection{Literature Review}
Econometric analysis of rank data builds on the random utility framework: under i.i.d.\ Type-I extreme value errors, a complete ranking admits a rank-ordered (exploded) logit likelihood that factors into a sequence of multinomial logits (\citealp{beggs1981assessing, chapaaan1982exploiting, hausman1987specifying}). These rank-based likelihoods have been used across applied preference measurement settings, ranging from parametric unfolding and multichoice logit models for incomplete rankings to stated-preference designs that translate ordered judgments into preference and welfare parameters (\citealp{desarbo1997parametric, ophem1999multichoice, calfee2001econometric}). More recently, platform search and merchandising environments generate ranking-implied signals through displayed positions and subsequent interactions, motivating structural estimation of preferences and ranking policies from exposure-driven data (\citealp{negahban2012iterative,compiani2024online}).
A central methodological theme is unobserved heterogeneity in ranking models: mixed-effects specifications introduce random coefficients in rank likelihoods (\citealp{bockenholt2001mixed}), finite mixtures and latent classes capture discrete heterogeneity in ranking patterns (\citealp{gormley2008mixture}), and semiparametric approaches relax distributional assumptions while maintaining the random-utility structure (\citealp{yan2019semiparametric}). Recent work further extends these ideas to richer covariate structures and modern data environments (\citealp{dong2025statistical}).

A complementary literature studies preference learning at scale using low-dimensional representations of unobserved tastes and product attributes. In econometrics and marketing, recent contributions combine revealed-preference inequalities with flexible regularization to enable counterfactual prediction in high-dimensional choice problems (\citealp{kallus2016revealed,donnelly2021counterfactual,armona2025learning,magnolfi2025triplet}). In parallel, the collaborative-filtering literature estimates user and item embeddings from implicit feedback using pairwise ranking objectives with a logistic form for utility differences, typically optimized via stochastic gradient methods (\citealp{rendle2011fast,oh2015collaboratively,he2016vbpr}). While often presented as algorithmic, these approaches are conceptually compatible with random utility models based on pairwise comparisons. Our paper integrates these strands by embedding a low-rank latent factor structure within an econometric logit likelihood for observed (and incomplete) rankings, where the set of observed comparisons can itself be selected because exposure depends on platform ordering, product popularity, or prior personalization. We address this endogenous observability using inverse probability weighting (IPW), reweighting each observed comparison by its estimated exposure propensity to recover preference parameters under standard selection-on-observables conditions. This correction parallels propensity-weighted debiasing in recommendation and learning-to-rank, which treats displayed rankings as treatments and uses inverse-propensity weights to obtain unbiased learning from biased feedback (\citealp{schnabel2016recommendations, joachims2017unbiased}). 

\section{Method}

\subsection{Observed Rankings} 
We assume that researchers observe rankings over a set of items \(j = 1,2,\ldots,m\) for individuals \(i = 1,2,\ldots,n\). Let \(\succeq_i\) denote the (latent) preference relation of individual \(i\), where \(j_1 \succeq_i j_2\) indicates that individual \(i\) prefers item \(j_1\) to item \(j_2\). Importantly, we allow the observed ranking information to be incomplete: for each individual, preferences are only observed for a subset of item pairs. 

It is useful to visualize the observed ranking information for an individual as a matrix. For illustration, consider the case with \(m=5\) items. The observed rankings for individual \(i\) can be represented by the following upper-triangular matrix $\mathbf{R}_i$:
\[
\mathbf{R}_i =
\begin{pmatrix}
- & \succeq & \cdot & \succeq & \cdot \\
  & -       & \succeq & \cdot & \cdot \\
  &         & -       & \cdot & \succeq \\
  &         &         & -     & \cdot \\
  &         &         &       & -
\end{pmatrix},
\]
where an entry \(\succeq\) in position \((j_1,j_2)\) indicates that the comparison \(j_1 \succeq_i j_2\) is observed, and \(\cdot\) denotes that the ranking between the two items is unobserved. Diagonal elements are omitted since self-comparisons are not meaningful.

Formally, let \(\mathcal{D}_i = \{(j_1,j_2) \mid j_1 \succeq_i j_2\}\) denote the set of observed pairwise rankings for individual \(i\). We do not require the same items, or the same item pairs, to be ranked by all individuals; consequently, the intersection \(\cap_{i=1}^n \mathcal{D}_i\) may be empty. Following standard assumptions, we assume that each individual’s observed preference relation \(\succeq_i\) is transitive, while allowing it to be incomplete because only a subset of pairwise comparisons is observed for each individual. Let \(\mathcal{D} = \{(i,j,j') \mid (j,j') \in \mathcal{D}_i\}\) denote the collection of all observed rankings across individuals. This set constitutes the data used for estimation, and our primary object of interest is to infer the unobserved preferences \((i,j,j') \notin \mathcal{D}\) that are not directly observed in the data.

In practice, observed rankings may arise from several sources. One common source is survey data in which individuals are asked to rank or compare a subset of items; however, eliciting complete rankings is often infeasible when the choice set is large. Rankings may also be constructed from rating data by converting numerical ratings into ordinal comparisons, though such rankings are typically incomplete because individuals do not rate all available items. A third and particularly important source is revealed preference inferred from observed choices. For example, if an individual consumes certain items (e.g., watches a movie on Netflix) but not others, this behavior can be interpreted as revealing a preference for the consumed items over those not chosen. In our empirical application, we construct the ranking data \(\mathcal{D}\) from transaction records provided by an online wine retailer. Specifically, we interpret wine \(A\) as being preferred to wine \(B\) for a given consumer if the consumer has purchased wine \(A\) but has not purchased wine \(B\) during the observation period.

\subsection{Modeling Utility Function}
Following the random utility framework \citep{mcfadden1973conditional}, we assume that the observed rankings are generated from the latent utility
\[
u_{ij} = x_j^\top \bm{\beta}_i + \alpha_j + \lambda_i^\top f_j + \varepsilon_{ij},
\]
where \(x_j\) is a vector of observed product attributes, \(\alpha_j\) is an item fixed effect capturing global popularity or average perceived quality, and \(\varepsilon_{ij}\) is an idiosyncratic utility shock that reflects unobserved factors affecting individual \(i\)’s evaluation of item \(j\). We allow preferences over observable attributes to be heterogeneous across individuals by permitting the coefficient vector \(\bm{\beta}_i\) to vary at the individual level. 

In addition to observable attributes, the utility specification includes a latent factor \(f_j \in \mathbb{R}^r\) that captures unobserved characteristics of item \(j\), where the dimension \(r\) is chosen by the researcher. Depending on the application, these latent characteristics may represent aspects such as overall style, quality gradients, usage occasions, or other dimensions of differentiation that are not directly measured in the data. In our empirical application using wine transactions, \(f_j\) may capture unobserved attributes such as the prestige associated with certain regions or stylistic features related to the taste of the wine that are not fully summarized by observed labels. The corresponding factor loadings \(\lambda_i\) are individual-specific, allowing consumers to differ in how they value these latent characteristics.

The inclusion of latent factors allows the model to capture systematic dependence in utility across individuals and items beyond what is explained by observed attributes and fixed effects. Items that tend to be ranked similarly by many consumers will acquire similar latent representations, while consumers who exhibit comparable ranking patterns will load similarly on these latent dimensions. As a result, information about preferences can be shared across individuals and across items, enabling the model to infer unobserved comparisons by exploiting common structure in ranking behavior rather than treating each consumer–item evaluation in isolation.



We estimate the model by interpreting the observed rankings \(\mathcal{D}\) as a collection of binary choice problems. Specifically, when we observe that individual \(i\) ranks item \(j\) above item \(j'\), we interpret this observation as revealing that, if both items were available simultaneously, individual \(i\) would choose item \(j\) over item \(j'\). Let
\[
\theta = \{\bm{\beta}_i, \alpha_j, \lambda_i, f_j\}_{i=1,\ldots,n;\, j=1,\ldots,m}
\]
denote the set of model parameters, and assume that the idiosyncratic utility shocks \(\varepsilon_{ij}\) follow the type-I extreme value distribution. Under this assumption, the probability that individual \(i\) prefers item \(j\) to item \(j'\) takes the logistic form
\begin{align*}
P(j \succeq j' \mid \theta)
&= \sigma\!\left(u_{ij} - u_{ij'}\right) \\
&= \frac{1}{1 + \exp\!\left[-(u_{ij} - u_{ij'})\right]},
\end{align*}
where \(\sigma(x) = \frac{1}{1+\exp(-x)}\) denotes the sigmoid function. 

Given the pairwise-choice interpretation, each observed comparison \((i,j,j') \in \mathcal{D}\) contributes a likelihood term equal to the probability that individual \(i\) prefers item \(j\) to item \(j'\). Under the logistic specification derived above, the likelihood contribution of a single observed pair is \(\sigma(u_{ij} - u_{ij'})\). Assuming independence of the idiosyncratic utility shocks across individuals and items, the log-likelihood of the full dataset \(\mathcal{D}\) is given by:
\[
\ell(\theta \mid \mathcal{D})
= \sum_{(i,j,j') \in \mathcal{D}} \ln \sigma\!\left(u_{ij} - u_{ij'}\right),
\]
which corresponds to the objective function used for estimation.



\subsection{Correcting for Selection in Observability}
The likelihood in Section 2.2 treats the observed comparison set $\mathcal{D}$ as if it were a representative sample of the underlying pairwise preference relation. In many ranking applications, however, which comparisons are observed is itself selected. We refer to this phenomenon as selection in observability: a pairwise comparison between items $j$ and $j'$ is recorded only when both items are simultaneously available to, encountered by, or considered by the individual.

Selection in observability is pervasive in practice. Survey respondents can rank only a limited number of items; in rating or consumption data, items are observed only if they are encountered; and in digital commerce, exposure is shaped by platform search, recommendations, assortment constraints, stockouts, and time-varying promotions. As a result, comparisons involving highly visible items are overrepresented, while comparisons involving niche or infrequently available items are systematically missing. Ignoring this selection can bias preference estimates by conflating exposure with taste. a concern that is closely related to consideration-set formation in marketing and exposure bias in recommender systems \citep{hauser1990evaluation,roberts1991development,joachims2017unbiased,schnabel2016recommendations}.

To formalize selection in observability, let $O_{ijj'} \in \{0,1\}$ denote an indicator that the comparison between items $j$ and $j'$ is \emph{observable} for individual $i$.
We interpret $O_{ijj'}=1$ as the event that both items enter the individual's effective consideration or availability set (e.g., both appear in the survey task or both are encountered through search and recommendation). The observed pairwise dataset $\mathcal{D}$ can be viewed as the collection of comparisons for which observability holds and an ordering is recorded:
\[
(i,j,j') \in \mathcal{D} \quad \Rightarrow \quad O_{ijj'}=1 \ \ \text{and we observe that } j \succeq_i j'.
\]
Let $\pi(x_j)\in(0,1)$ denote the (marginal) probability that item $j$ is observable as a function of item-side observables $x_j$. We assume that, conditional on observables, the probability that a pair is observable factorizes as
\begin{equation*}
\label{eq:pair_observability_factorization}
\pi_{jj'}=\Pr\!\left(O_{ijj'}=1 \mid x_j, x_{j'}\right) \;=\; \pi(x_j)\,\pi(x_{j'}) \;\equiv\; \pi_j\,\pi_{j'}.
\end{equation*}
This assumption captures the idea that observability is primarily driven by item-level exposure or availability, and implies that rarely available items mechanically generate fewer observed comparisons with any other item. As detailed in the next subsection, we will adopt inverse probability weighting (IPW) based on estimates of $\pi(\cdot)$ to adjust for selection in observability.

\subsection{Estimation and Computation}
Optimizing the likelihood function poses several technical challenges. First, the model is high-dimensional, as it includes both individual-specific parameters \((\bm{\beta}_i, \lambda_i)\) and item-specific parameters \((\alpha_j, f_j)\). Such high dimensionality can lead to numerical instability and noisy estimates, particularly in settings with sparse ranking data. Second, the cardinality of \(\mathcal{D}\) can be very large, since it contains all observed pairwise comparisons implied by the rankings. Directly optimizing the likelihood using the full dataset \(\mathcal{D}\) is therefore computationally demanding.

To address the first challenge, we introduce regularization into the estimation procedure. Let $\Theta$ denote
the parameter space. We estimate $\theta$ by minimizing an IPW-corrected penalized negative log-likelihood:
\[
\hat{\theta}
=
\arg\max_{\theta\in\Theta}
\sum_{(i,j,j')\in D}
\frac{1}{\hat{\pi}_{jj'}} \ln\sigma(u_{ij}-u_{ij'})
\;-\;
\kappa\|\theta\|_2^2
,
\]
where $\kappa>0$ is the penalty term and $\hat{\pi}_{jj'}$ is the estimate of the probability that the comparison between $j$ and $j'$ is observed. The inverse-probability weighting $\frac{1}{\hat{\pi}_{jj'}} $ corrects for selection by upweighting under-considered pairs \citep{horvitz1952generalization,rosenbaum1983central}. Rather than estimating a separate propensity for each unordered pair $\{j,j'\}$, which is infeasible when $m$ is large,
we exploit the factorization and impose a parsimonious parametric model on the marginal observability probability. Specifically, we parameterize
\[
\pi_j \equiv \pi(x_j;\psi) = \sigma(x_j^\top \psi),
\]
where $\psi$ is an auxiliary parameter vector (distinct from the preference parameters $\theta$) and $x_j$ includes an
intercept. This specification plays the role of a propensity-score model: it maps item-side observables into a
probability that item $j$ enters the consideration/observability process, and it induces the pairwise comparison
probability $\pi_{jj'}=\pi(x_j;\psi)\pi(x_{j'};\psi)$.

To estimate $\psi$ using based on which comparisons are observed, we construct, for every individual $i$ and unordered pair
$\{j,j'\}$, the observability indicator
\[
O_{ijj'} \;=\; \mathbf{1}\!\left\{(i,j,j')\in D\ \text{or}\ (i,j',j)\in D\right\}.
\]
Under the imposed model, $O_{ijj'}$ is a Bernoulli random variable with success probability
$\pi_{jj'}(\psi)=\pi(x_j;\psi)\pi(x_{j'};\psi)$. Let $N_{jj'}=\sum_{i=1}^n O_{ijj'}$ denote the number of individuals for
whom the unordered pair $\{j,j'\}$ is observed. The resulting (aggregated) log-likelihood for $\psi$ is
\[
\ell_O(\psi)
=
\sum_{1\le j<j'\le m}
\left[
N_{jj'} \log\!\big(\pi(x_j;\psi)\pi(x_{j'};\psi)\big)
+
(n-N_{jj'}) \log\!\Big(1-\pi(x_j;\psi)\pi(x_{j'};\psi)\Big)
\right],
\]
and we estimate $\psi$ by maximum likelihood,
\[
\hat{\psi}=\arg\max_{\psi}\ \ell_O(\psi).
\]
We then obtain plug-in estimates
\[
\hat{\pi}_j=\pi(x_j;\hat{\psi})=\sigma(x_j^\top \hat{\psi})
\quad\text{and}\quad
\hat{\pi}_{jj'}=\hat{\pi}_j\,\hat{\pi}_{j'},
\]
which we use in the inverse-probability weights in the main objective.

To address the second challenge, we employ stochastic gradient descent (SGD) to optimize the objective function.
For the IPW objective, we implement an inverse-probability resampling scheme: rather than sampling comparisons uniformly
from $D$, at each iteration we draw a single observed comparison $(i,j,j')\in \mathcal{D}$ from
\begin{equation*}
\label{eq:q_ipw_pair}
q(i,j,j')
\;=\;
\frac{\hat{\pi}_{jj'}^{-1}}{\sum_{(i',k,k')\in \mathcal{D}} \hat{\pi}_{kk'}^{-1}}.
\end{equation*}
Sampling from $q$ moves the inverse-probability correction into the data stream, so that the per-sample gradient can be
taken with respect to the unweighted log-likelihood term (with the normalizing constant absorbed into the learning rate)
\citep{zhao2015stochastic}. The resulting SGD update retains the same analytic form as in the unweighted case:
\begin{equation*}
\theta \leftarrow \theta + \eta
\left(
\frac{e^{-(u_{ij}-u_{ij'})}}{1 + e^{-(u_{ij}-u_{ij'})}}
\cdot
\frac{\partial (u_{ij}-u_{ij'})}{\partial \theta}
\;-\;
\kappa \theta
\right),
\end{equation*}
where $(i,j,j')$ is drawn from $\mathcal{D}$ and $\eta>0$ is the learning rate.

Computationally, the SGD algorithm can be implemented without scanning all elements of $\mathcal{D}$ because $\hat{\pi}_{jj'}$ depends
only on the pair $\{j,j'\}$. We precompute $\hat{\pi}_{jj'}$ for each observed pair and maintain, for each
pair, the list of indices in $D$ that correspond to that pair. We then (i) sample an unordered pair $\{j,j'\}$ using an
alias table with weights proportional to $\hat{\pi}_{jj'}^{-1}\,|\mathcal{D}_{jj'}|$, where $|\mathcal{D}_{jj'}|$ is the number of observed
comparisons in $D$ involving $\{j,j'\}$, and (ii) sample uniformly from the stored indices within $
\mathcal{D}_{jj'}$ to obtain a
triple $(i,j,j')$. This two-stage procedure yields exact draws for optimizing the objective function while preserving the per-iteration
cost of standard SGD.


\section{Empirical Illustration}
\subsection{Data and Background}
We illustrate the proposed preference-learning framework using proprietary transaction data from a leading online alcoholic-beverage retailer in Taiwan, covering all orders placed on the platform between 2021 and 2024. While the retailer offers a broad portfolio of alcoholic products—including sake and whisky—wine constitutes its core category in terms of both sales volume and product variety. To maintain a focused empirical setting, we therefore restrict attention to wine purchases throughout the analysis.

The dataset comprises 311,089 transaction records from 23,721 unique customers. Each transaction is linked to a persistent customer identifier, allowing us to reconstruct individual purchase histories over time. For every order, we observe the transaction timestamp, a unique product identifier, quantity purchased, and price paid, along with detailed product attributes such as country of origin, region, color, sweetness level, vintage, and bottle size. In addition, self-reported demographic information—including age and gender—is available for a subset of customers. Taken together, these data form a rich panel of consumer–product interactions: the detailed attribute information supports a structured representation of wine characteristics, while the longitudinal purchase histories allow us to infer relative preferences from revealed choice behavior. This combination makes the setting particularly well suited for studying heterogeneous preferences using ranking-based methods.

Table~\ref{table:summarystats_customer} reports customer-level summary statistics for all customers who made at least one purchase during the 2021--2024 period. The customer base is slightly male-skewed (59.7\%) with an average age of approximately 40. 
Annualized spending averages 10{,}322~NTD (USD~323), but the distribution is highly right-skewed (SD 37{,}945~NTD; USD~1{,}186), indicating a small group of heavy buyers. Customers purchase relatively infrequently—2.65 orders per year on average—implying each transaction is sizable, consistent with occasional, high-involvement purchases. In terms of product composition, preferences tilt strongly toward Old World wines: 63\% of customers purchase France at least once, and 84.7\% purchase from Old World regions overall.\footnote{In the wine context, \emph{Old World} refers to traditional European wine-producing regions such as France, Italy, Spain, and Germany, while \emph{New World} refers to producers outside Europe, including the United States, Chile, Australia, New Zealand, and South Africa.} Nevertheless, New World wines are also common, with 50\% of customers purchasing at least one bottle. Finally, participation is concentrated in the low and mid price tiers—defined as 500--1{,}000~NTD (USD~16--31) and 1{,}001--3{,}000~NTD (USD~31--94), respectively—with 61\% and 66\% of customers purchasing at least once in these ranges. By contrast, only 9\% of customers ever purchase high-end wines, defined as prices above 10{,}000~NTD (USD~313+). These patterns motivate a modeling approach that allows for substantial heterogeneity in both origin preferences and price sensitivity.

\renewcommand{\arraystretch}{1.5} 
\begin{table}[!htbp] 
\centering 
\begin{threeparttable} 
\caption{Customer Summary Statistics} \label{table:summarystats_customer} 
\begin{tabular}{lcccc} 
\hline \hline 
& \multicolumn{1}{c}{Average} & \multicolumn{1}{c}{SD} & \multicolumn{1}{c}{Min} & \multicolumn{1}{c}{Max} \\ \hline \textit{A. Demographics} & & & & \\ 
Female & 0.428 & 0.495 & 0 & 1\\
Age (years) & 39.93 & 11.09 & 18 & 83\\\midrule 
\textit{B. Purchase Behavior (Annualized)} & & & &\\
Spending (NTD) & 10{,}322 & 37{,}945 & 166 & 1{,}120{,}101 \\
Spending (USD) & 323 & 1{,}186 & 5.1 & 35{,}003 \\ 
Purchase Frequency & 2.65 & 7.41 & 0.25 & 158.86 \\
\midrule 
\multicolumn{5}{l}{\textit{C. Region ever Purchased}}\\
France & 0.63 & 0.48 & 0 & 1 \\
Italy & 0.38 & 0.48 & 0 & 1 \\
US & 0.22 & 0.41 & 0 & 1 \\
Old World & 0.847 & 0.36 & 0 & 1\\
New World & 0.50 & 0.5 & 0 & 1 \\
\midrule
\multicolumn{5}{l}{\textit{D. Wine Type ever Purchased }}\\
Red Wine & 0.71 & 0.45 & 0 & 1 \\\
White Wine & 0.50 & 0.5 & 0 & 1 \\
Sparkling & 0.31 & 0.46 & 0 & 1 \\ 
\midrule
\multicolumn{5}{l}{\textit{E. Price Tier ever Purchased (in USD)}}\\
Low Price ($\leq$ 31) & 0.61 & 0.49 & 0 & 1 \\
Mid Price (31 -- 94)& 0.66 & 0.47 & 0 & 1 \\ 
Mid-High Price (94 -- 313) &  0.28 & 0.44 & 0 & 1 \\ 
High Price ($\ge$ 313)& 0.09 & 0.29 & 0 & 1 \\ 
\hline 
\end{tabular}
\begin{tablenotes}[flushleft] \small
\item \textit{Note:} The unit of observation is customer. The sample consists of $23{,}721$ observations; 
Panel~A reports demographics for customers who made at least one purchase during the 2021--2024 sales period. Panel~B reports annualized purchase behavior, where spending, purchase frequency,  and items purchased are averaged across the four-year sample period. 
Panel~C--E reports purchase propensities based on indicator variables for whether a customer purchased at least one bottle from the corresponding origin group, wine type, or price tier during the sample period, whose indicate the share of customers with the indicator equal to one. 
\end{tablenotes} 
\end{threeparttable} 
\end{table}





\subsection{Construction of Ranking Data}

To address the extreme sparsity inherent in transaction-level wine data, we aggregate individual stock-keeping units (SKUs) into economically meaningful wine categories. This aggregation is designed to reduce dimensionality while preserving the attributes most relevant for consumer differentiation. The resulting representation reflects extensive consultation with domain experts from the retailer and balances statistical tractability with interpretability.

We begin by standardizing raw product information to obtain a consistent description of each wine. The catalog reports a four-level geographic hierarchy (country, major region, appellation, and sub-appellation), and wine color (e.g., red, white). To mitigate sparsity at fine geographic levels, we collapse the hierarchy to country (e.g., France) and major region (e.g., Bordeaux) , and drop the more granular sub-appellation level (e.g., the Pauillac village in Bordeaux). This choice retains the primary origin distinctions that consumers commonly use in evaluating wines, while avoiding categories that are too sparsely populated for reliable estimation. 

Next, we discretize prices into tiers intended to capture meaningful differences in consumer spending behavior rather than arbitrary numeric cutoffs. Guided by the retailer’s domain knowledge and industry experience,, we define the following price tiers: up to 500 NTD; 501--1{,}000; 1{,}001--2{,}000; 2{,}001--3{,}000; 3{,}001--4{,}000; 4{,}001--5{,}000; 5{,}001--10{,}000; 10{,}001--20{,}000; and above 20{,}000 New Taiwanese Dollar (NTD).\footnote{Using an exchange rate of 1 USD = 32 NTD, these cutoffs correspond to approximately USD 16, 31, 63, 94, 125, 156, 313, 625, and above USD 625.} This discretization preserves vertical differentiation in price while reducing noise from idiosyncratic pricing and infrequently purchased premium SKUs.

In parallel, we standardize wine styles using a rule-based taxonomy designed to reduce heterogeneity arising from inconsistent grape-variety nomenclature and overly granular blend definitions. Grape variety names are first standardized by merging synonyms and region-specific aliases, after which each wine is assigned an initial style label based on the set of grape varieties used, abstracting from blending proportions. Within each region, styles are ranked by product prevalence, with those collectively accounting for the top 80\% of products classified as major styles; remaining low-frequency styles are treated as minor and collapsed into broader ``red wine'' or ``white wine'' categories. Major styles  are then mapped to a set of canonical global wine styles (e.g., the ``Bordeaux Blend'' style that mainly composes of Cabernet Sauvignon and Merlot) using predefined rules linking grape-variety combinations to widely recognized style labels. Wines that do not match any canonical rule retain their original variety-based labels. A detailed description of the taxonomy and mapping rules is provided in Appendix Section~6.1. We then combine origin (country--region--appellation), grape variety, and price tier to define a composite product segment for each wine. For example, a red wine from France, Bordeaux, 
Pauillac priced at 2{,}300 NTD is classified as ``France--Bordeaux, Bordeaux Blend, 2{,}001--3{,}000.'' 
In all subsequent analysis, these composite product segments are treated as the  ``items'' in the model. 

Finally, we construct a set of observed pairwise rankings that encode revealed preferences at the customer level as the following. For a given customer $i$, item $j$ is said to be preferred to item $j'$ if the customer purchased at least one wine belonging to item $j$ during the sample period but did not purchase any wine belonging to item $j'$. Each such comparison is interpreted as a revealed preference statement that $j \succ_u j'$. Formally, the dataset $\mathcal{D}$ consists of all observed triplets $(i, j, j')$ satisfying this condition. The collection $\mathcal{D}$ therefore aggregates, for each customer, a set of pairwise comparisons between purchased and unpurchased items. These comparisons form an incomplete but informative ranking over the item space, reflecting relative preferences inferred from observed purchase behavior. 




\subsection{Model Specificaiton}
Because each item $j$ is defined as a composite of region ($r$), grape variety ($g$), and price tier ($p$), it is convenient to index items by the triple $(r,g,p)$. We denote the latent utility that individual $i$ derives from item $(r,g,p)$ by $u_{irgp}$. To capture preferences over observed product characteristics while preserving the collaborative-filtering advantages of the method, we specify the utility function as
\begin{equation}
\label{eq:utility_decomposition}
u_{irgp}
= \delta_{r}
+ \gamma_{g}
+ \pi_{p}
+ \lambda_i^{\top}{f}_{rgp}
+ \varepsilon_{irgp},
\end{equation} 
where $\delta_r$ is a region fixed effect, $\gamma_g$ is a grape-variety fixed effect, and $\pi_p$ is a price-tier fixed effect. These components jointly summarize the contribution of observed product attributes to utility. The idiosyncratic error term $\varepsilon_{irgp}$ captures unobserved taste shocks and is assumed to follow a type-I extreme value distribution.

The term $ \lambda_i^{\top}{f}_{rgp}$ represents the latent factor component of the model. Here $\lambda_i$ is an individual-specific preference vector and ${f}_{r,g,p}$ is a latent feature vector associated with item $(r,g,p)$. These latent dimensions capture aspects of product valuation that are not directly observed in the data but systematically influence choice. In the wine context, they may reflect preferences over abstract attributes such as perceived prestige, taste, or a preference for certain flavor profiles that cut across formal grape or price classifications. They may also absorb context-specific demand factors, such as whether purchases are intended for gifting or celebration versus routine consumption. By conditioning on observed attributes (region, grape variety, and price tier) and augmenting them with a flexible latent factor structure, the model accommodates rich heterogeneity in preferences while preserving interpretability along economically meaningful dimensions.

The factor structure also enables the model to learn similarity patterns from co-purchase behavior. Consumers who exhibit similar ranking patterns over observed items are placed nearby in the latent preference space, while items that tend to be purchased by similar sets of consumers acquire similar latent representations. This data-driven embedding allows the model to extrapolate preferences to unobserved items and to capture substitution patterns that are not fully explained by observable product characteristics alone.


\subsection{Estimation Results}

\begin{figure*}[!htbp]
\centering

\begin{subfigure}[t]{0.32\textwidth}
  \centering
  \includegraphics[width=\linewidth]{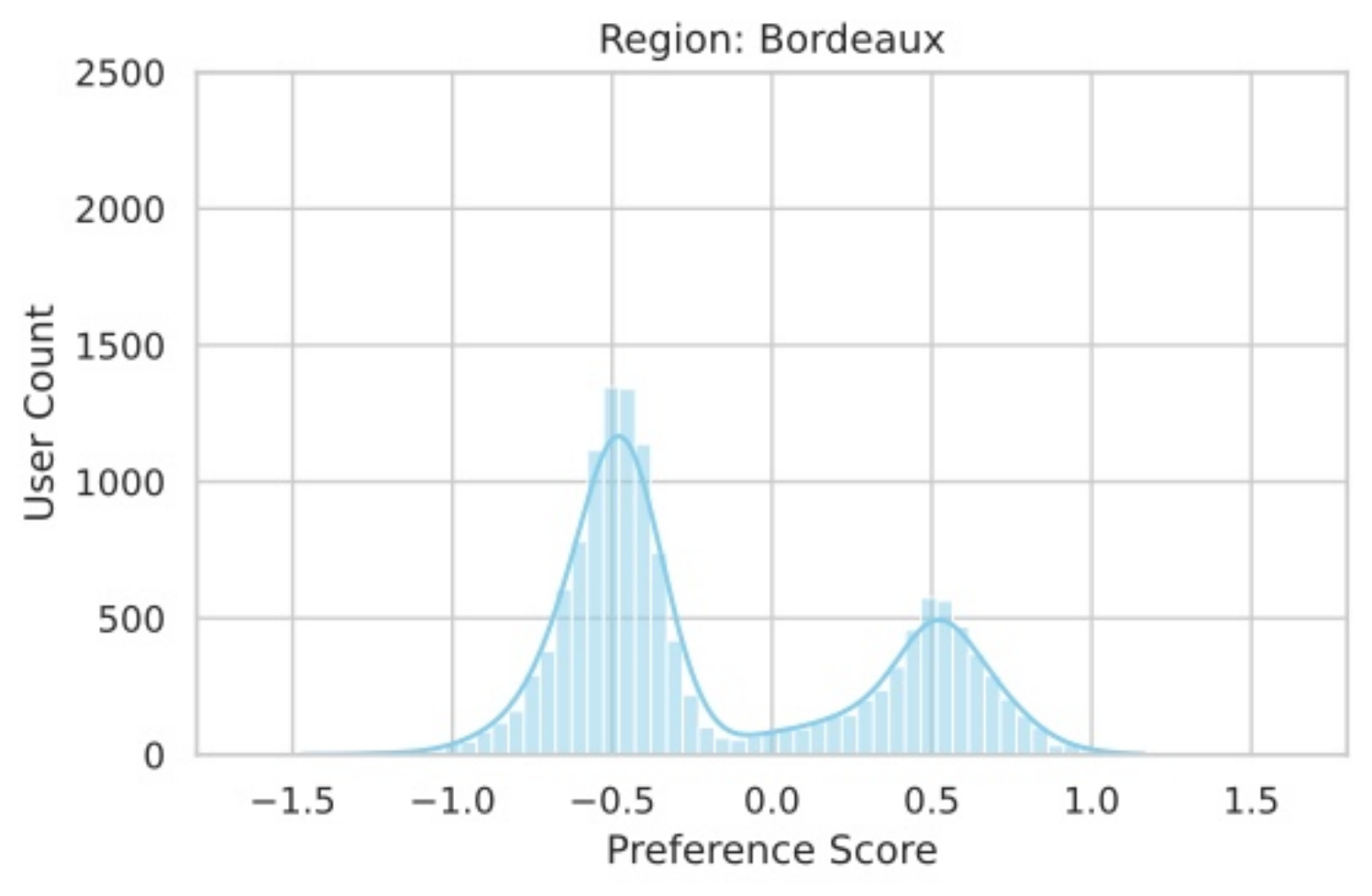}
  \caption{Bordeaux}
\end{subfigure}\hfill
\begin{subfigure}[t]{0.32\textwidth}
  \centering
  \includegraphics[width=\linewidth]{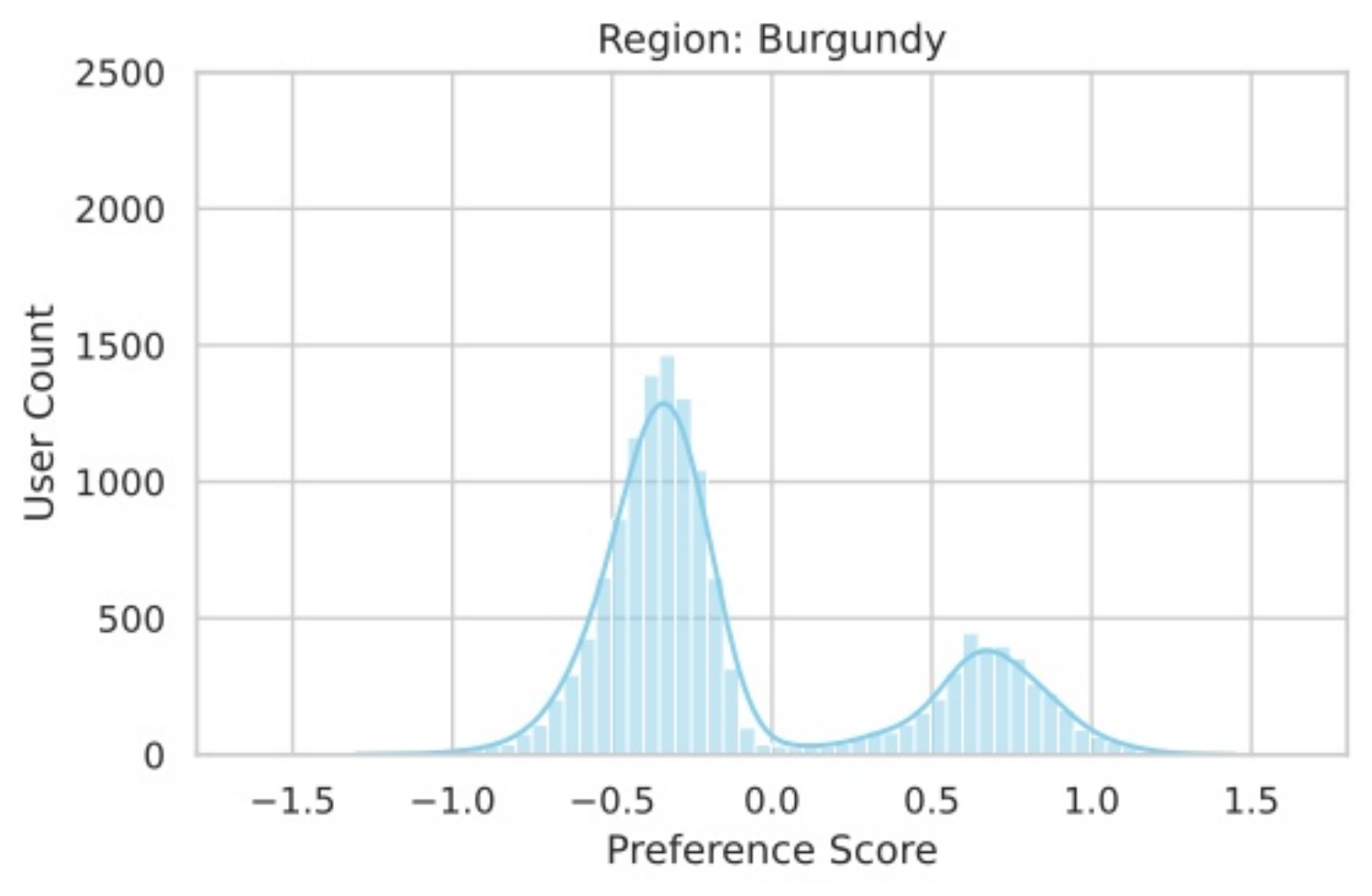}
  \caption{Burgundy}
\end{subfigure}\hfill
\begin{subfigure}[t]{0.32\textwidth}
  \centering
  \includegraphics[width=\linewidth]{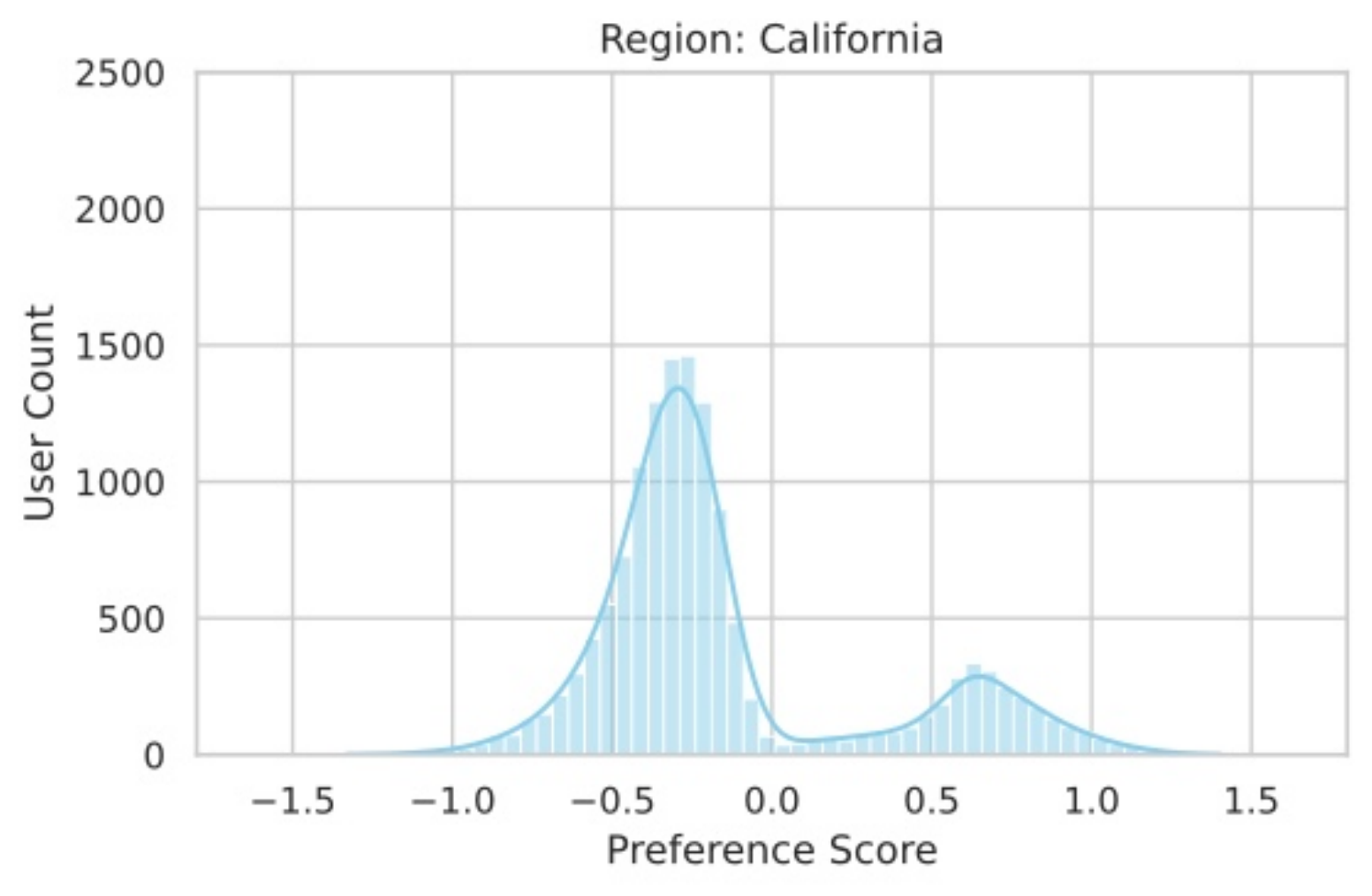}
  \caption{California}
\end{subfigure}

\vspace{0.6em}

\begin{subfigure}[t]{0.32\textwidth}
  \centering
  \includegraphics[width=\linewidth]{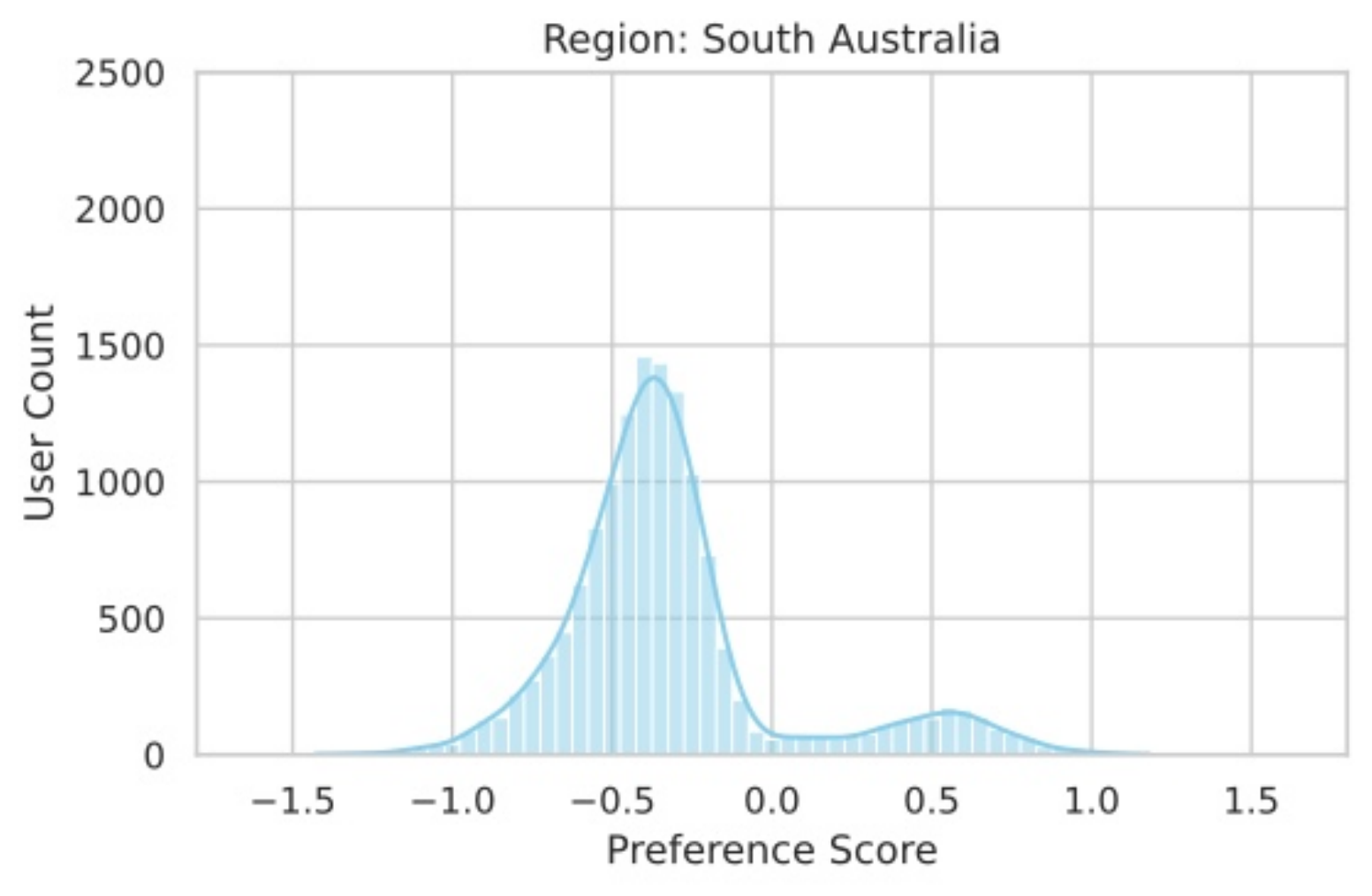}
  \caption{South Australia}
\end{subfigure}\hfill
\begin{subfigure}[t]{0.32\textwidth}
  \centering
  \includegraphics[width=\linewidth]{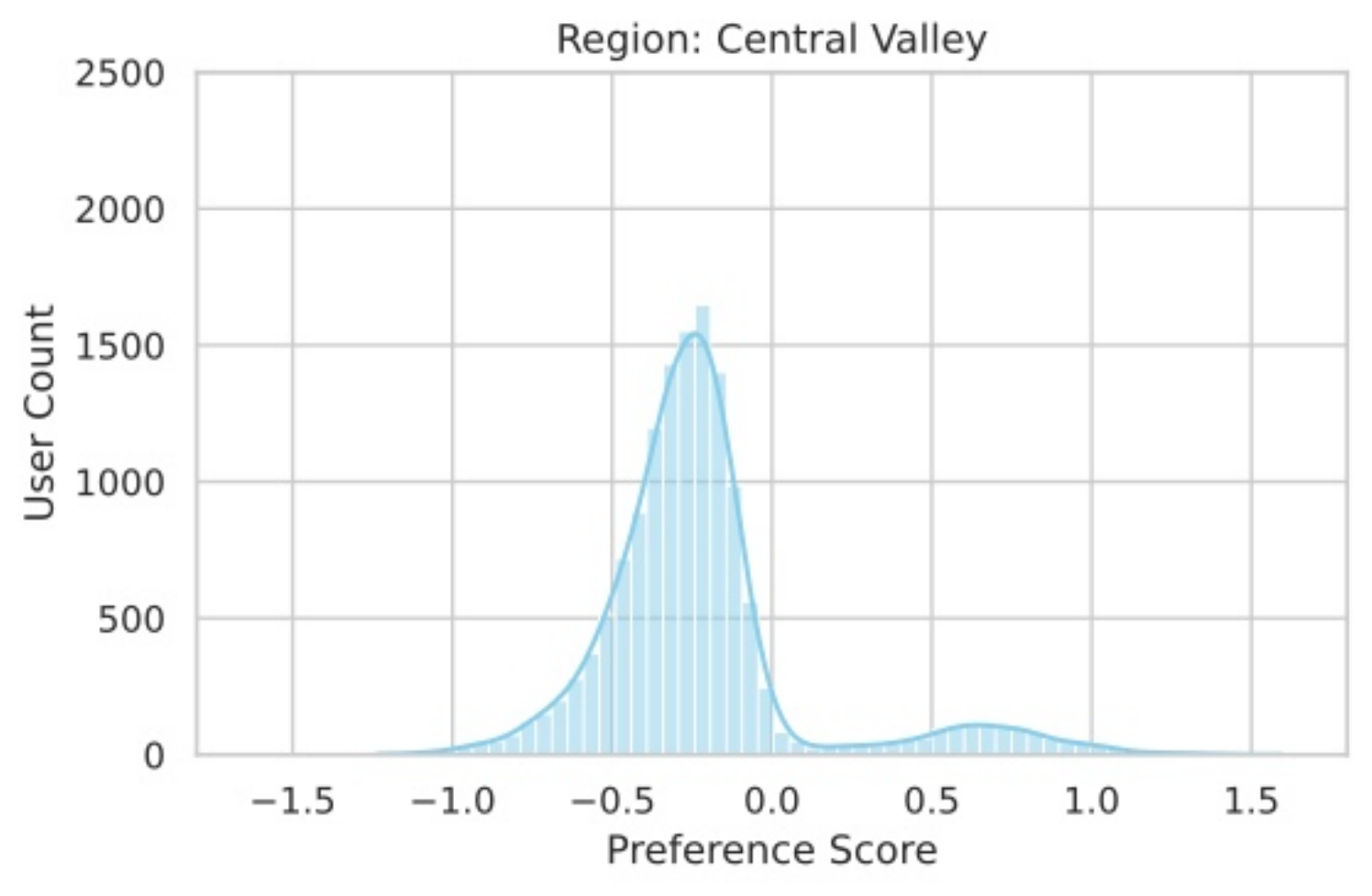}
  \caption{Central Valley}
\end{subfigure}\hfill
\begin{subfigure}[t]{0.32\textwidth}
  \centering
  \includegraphics[width=\linewidth]{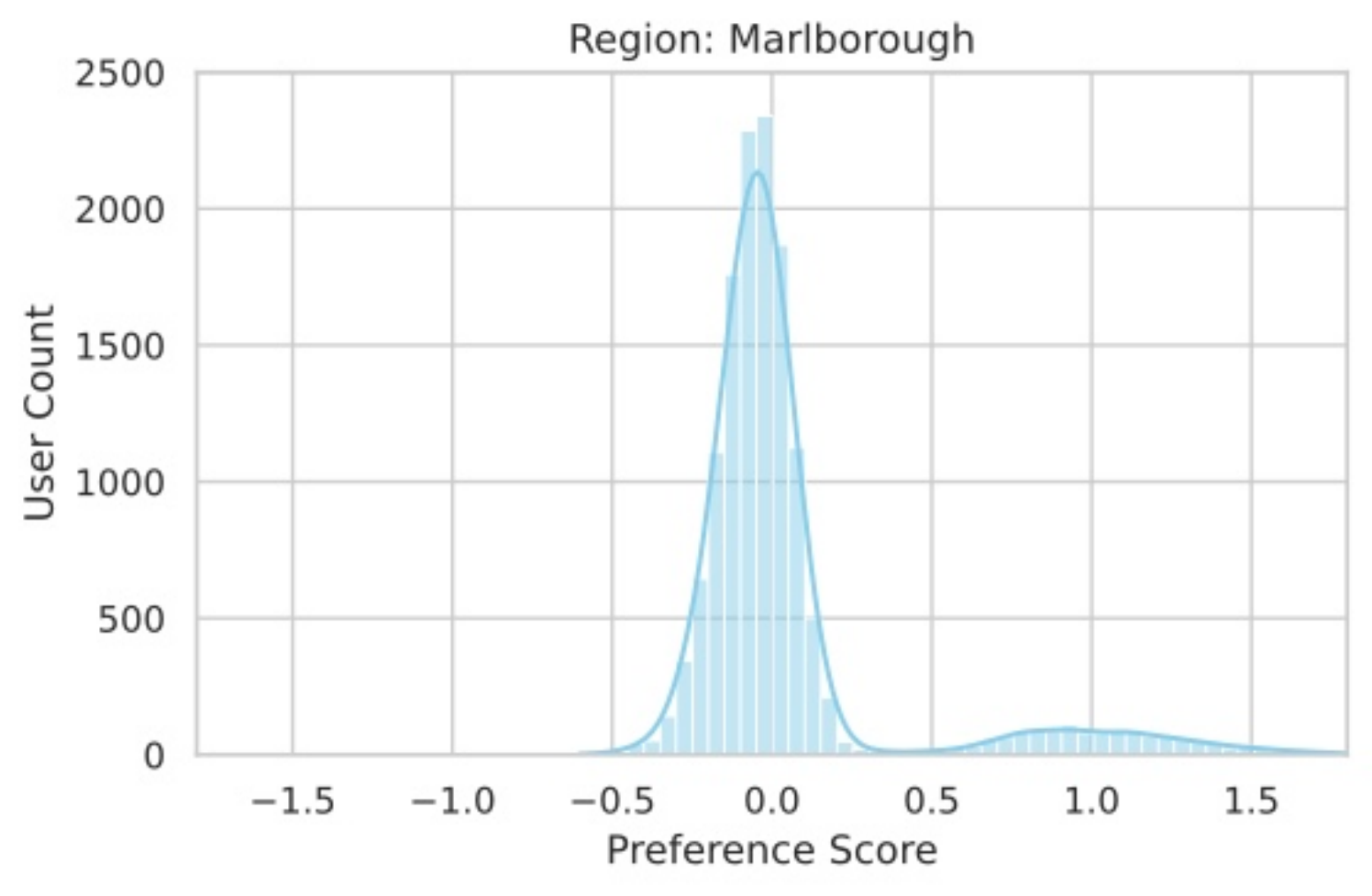}
  \caption{Marlborough}
\end{subfigure}

\caption{Distributions of individual-specific region effect coefficients $\delta_r$ for the six regions.}
\label{fig:region_coef_dists}

\end{figure*}

We begin by summarizing the distribution of estimated region-specific preference effects across consumers. To provide a compact and interpretable view of preference heterogeneity, we focus on six representative wine regions that span both the Old World and the New World: Bordeaux and Burgundy (France), California (United States) and Central Valley (Chile), South Australia (Australia), and Marlborough (New Zealand). For each region, Figure~\ref{fig:region_coef_dists} plots the empirical distribution of the individual-specific region coefficients \( \delta_{r,i} \), which capture how strongly a given consumer ranks wines from region \( r \) relative to other regions, holding grape variety and price tier fixed.

In Figure \ref{fig:region_coef_dists}, several patterns emerge. First, historically prestigious and widely consumed regions such as Bordeaux and Burgundy exhibit distributions that are shifted rightward relative to other regions, indicating higher average preference levels and a substantial mass of consumers with strongly positive region-specific utility. At the same time, these distributions are clearly dispersed and often bimodal, revealing pronounced heterogeneity. In contrast, regions with more specialized or stylistically distinctive profiles—such as Marlborough—display distributions that are more tightly concentrated around zero with thinner right tails, suggesting a narrower but more homogeneous appeal. Taken together, these results indicate that regional origin operates not merely as a mean-shifting attribute but as a salient dimension of preference heterogeneity, underscoring the importance of allowing region effects to vary flexibly across individuals.

\begin{figure}[!htbp]
  \centering
  \includegraphics[width=0.95\textwidth]{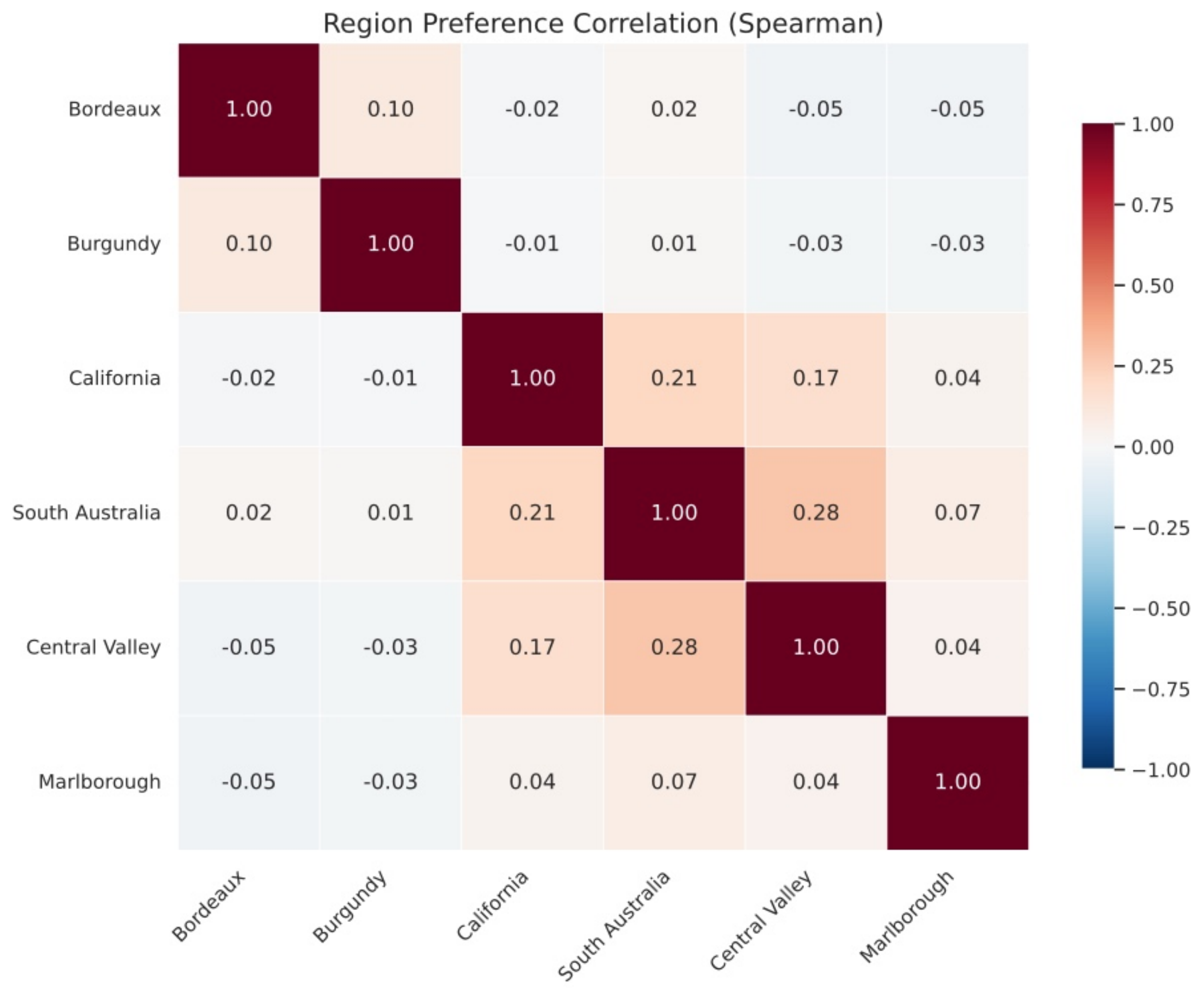}
  \caption{Correlation matrix of customer-specific region preference effects. Each entry reports the pairwise correlation between two regions' preference effects.}
  \label{fig:region_corr}
\end{figure}

We inspect correlation between regional preference in Figure~\ref{fig:region_corr}. Two patterns stand out. First, the three New World regions (California, South Australia, and Central Valley) exhibit the strongest positive comovement in preferences. This clustering suggests that relative affinity for one of these regions also tend to display higher affinity for the others. Second, the two French benchmark regions (Bordeaux and Burgundy) are positively correlated, but only modestly ($0.10$), indicating that ``liking France'' is not a dominant single dimension of heterogeneity once the model accounts for other product characteristics. Moreover, the correlations between the French regions and some New World regions are slightly negative (e.g., Bordeaux--Central Valley: $-0.05$), consistent with mild substitution in rankings between these styles for some customers. 


Finally, Marlborough, which is most known for Sauvignon white wines, appears relatively distinct: its correlations with the New World cluster are near zero and remain small even with South Australia. 

\begin{figure*}[!htbp]
\centering

\begin{subfigure}[t]{0.32\textwidth}
  \centering
  \includegraphics[width=\linewidth]{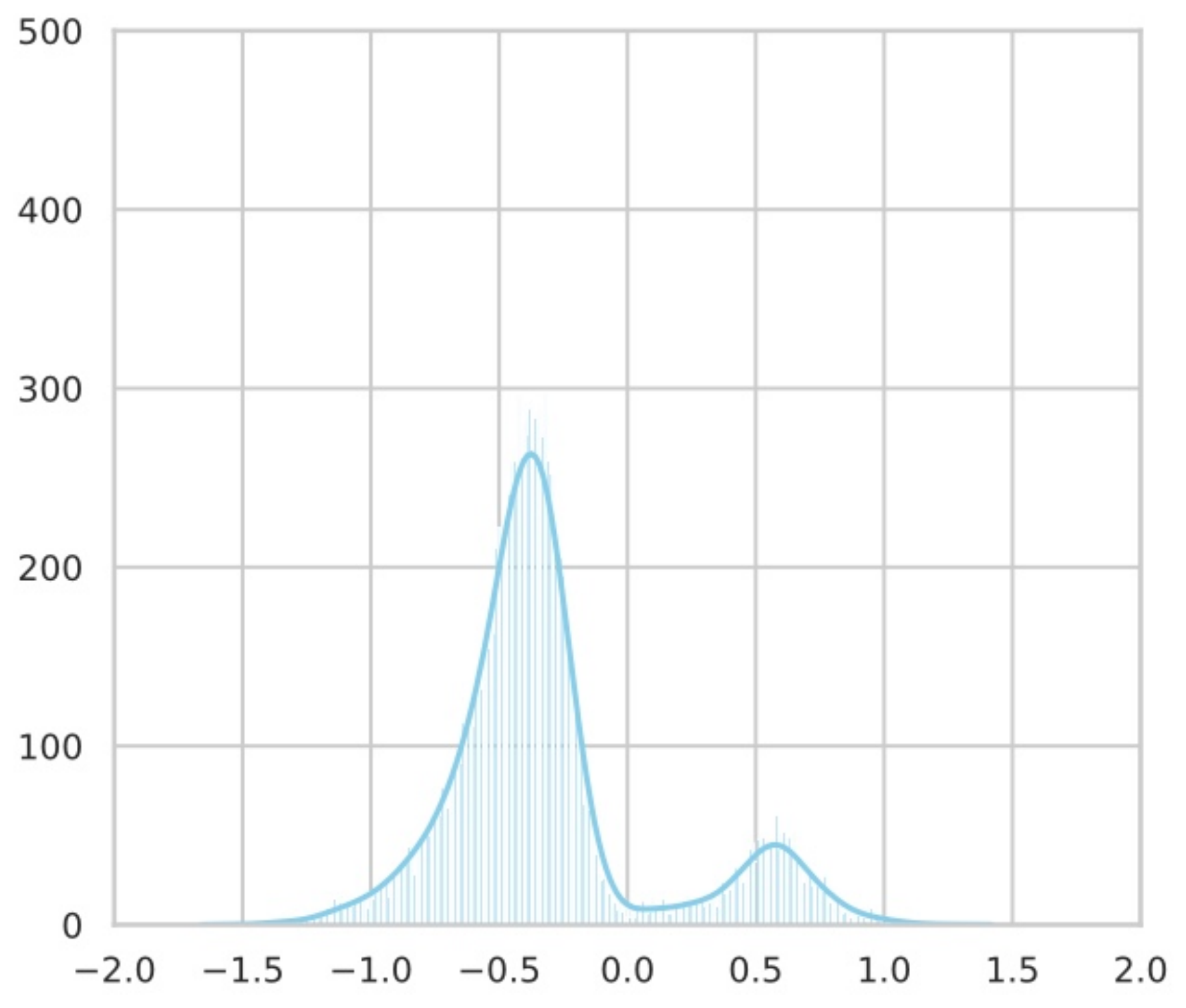}
  \caption{NTD $\le$ 500}
\end{subfigure}\hfill
\begin{subfigure}[t]{0.32\textwidth}
  \centering
  \includegraphics[width=\linewidth]{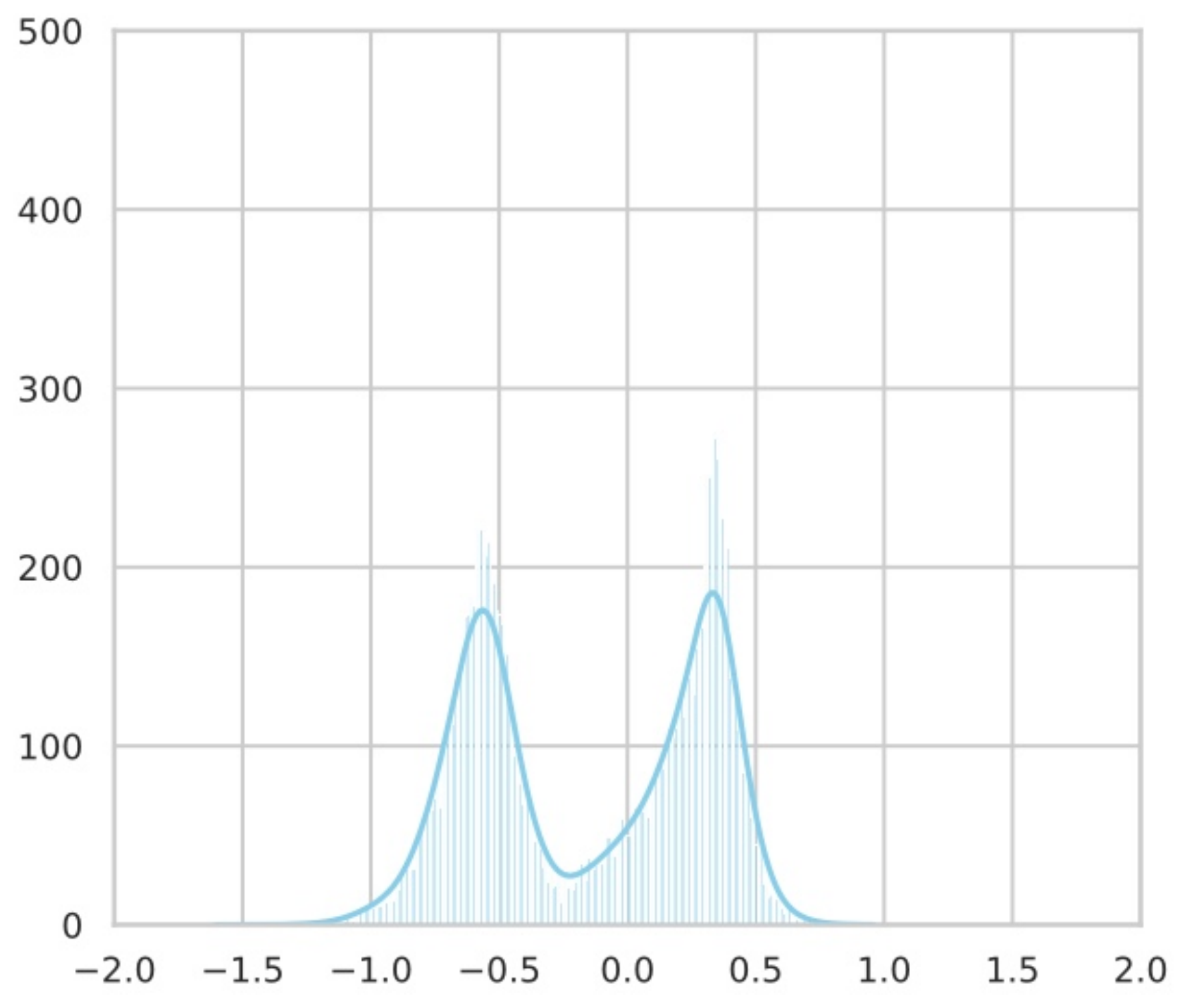}
  \caption{NTD 501--1,000}
\end{subfigure}\hfill
\begin{subfigure}[t]{0.32\textwidth}
  \centering
  \includegraphics[width=\linewidth]{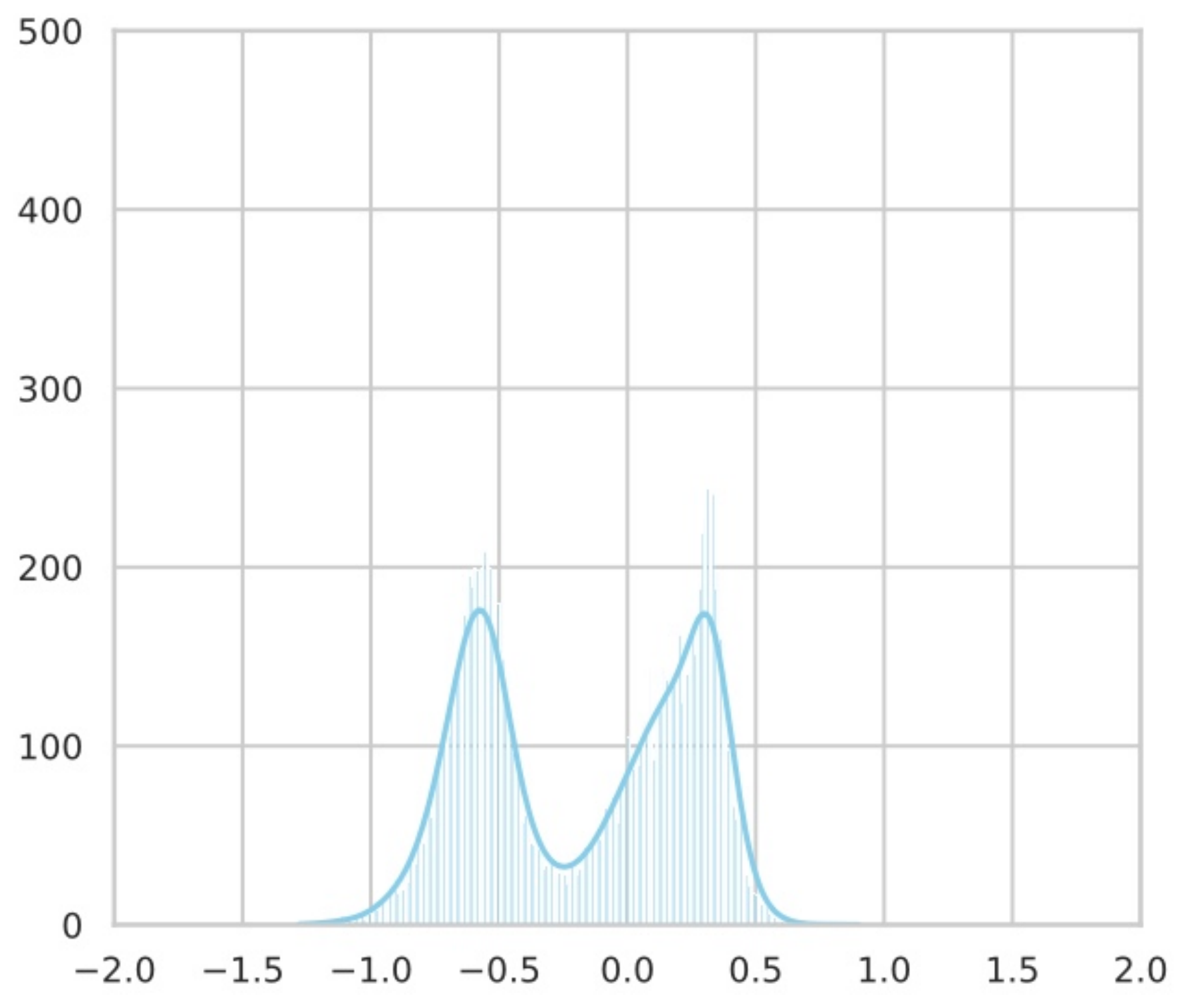}
  \caption{NTD 1,001--2,000}
\end{subfigure}

\vspace{0.6em}

\begin{subfigure}[t]{0.32\textwidth}
  \centering
  \includegraphics[width=\linewidth]{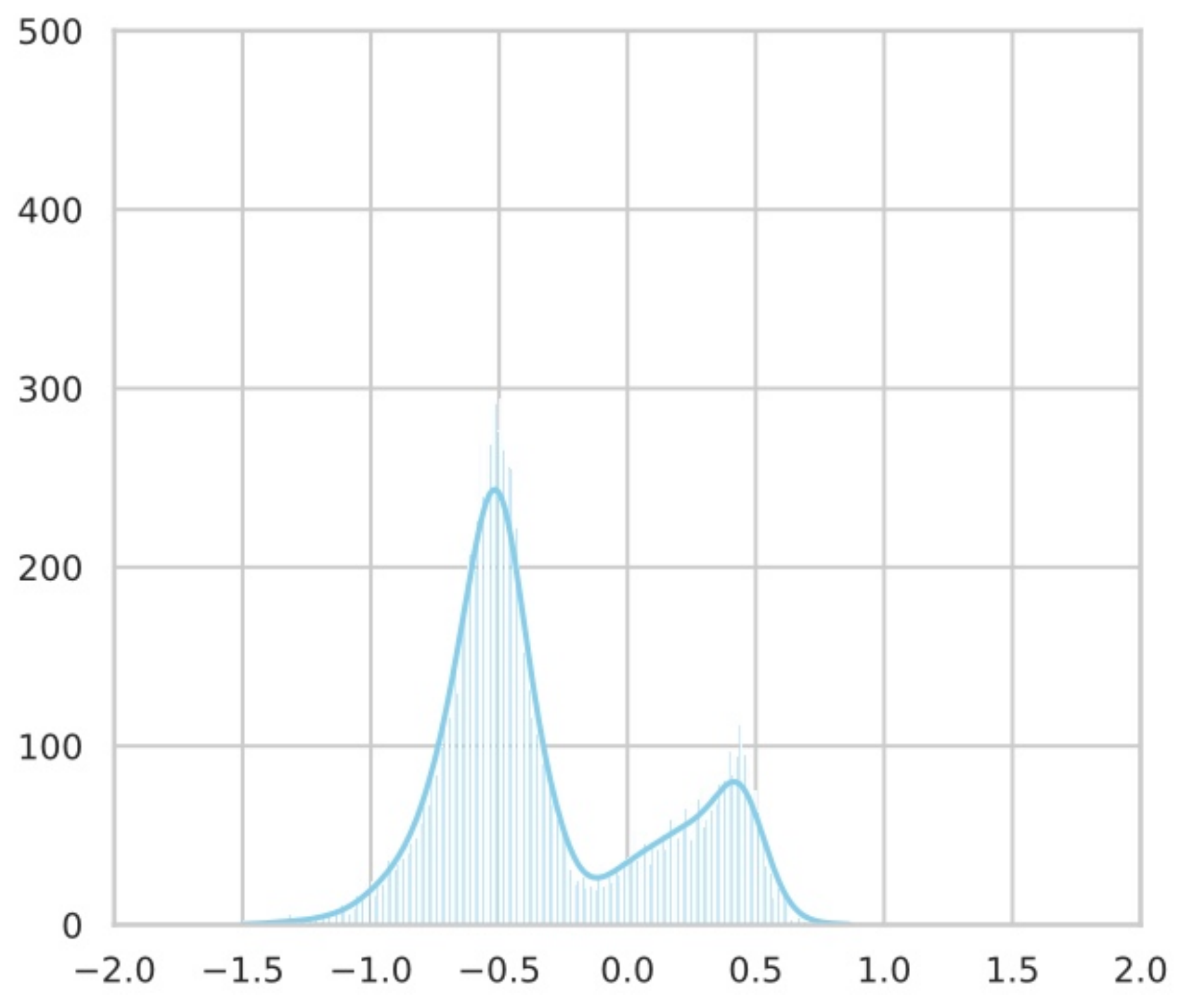}
  \caption{NTD 2,001--3,000}
\end{subfigure}\hfill
\begin{subfigure}[t]{0.32\textwidth}
  \centering
  \includegraphics[width=\linewidth]{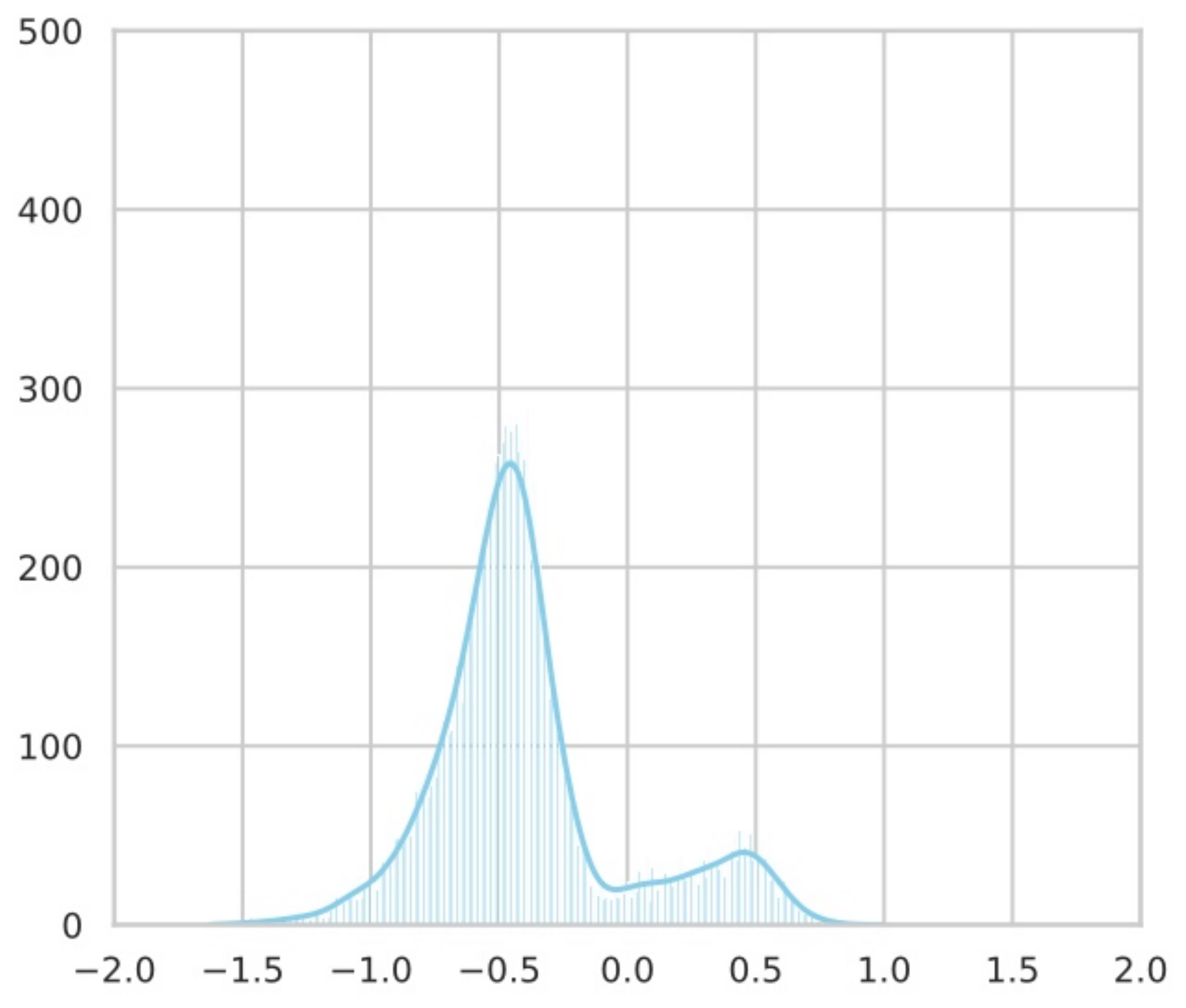}
  \caption{NTD 3,001--4,000}
\end{subfigure}\hfill
\begin{subfigure}[t]{0.32\textwidth}
  \centering
  \includegraphics[width=\linewidth]{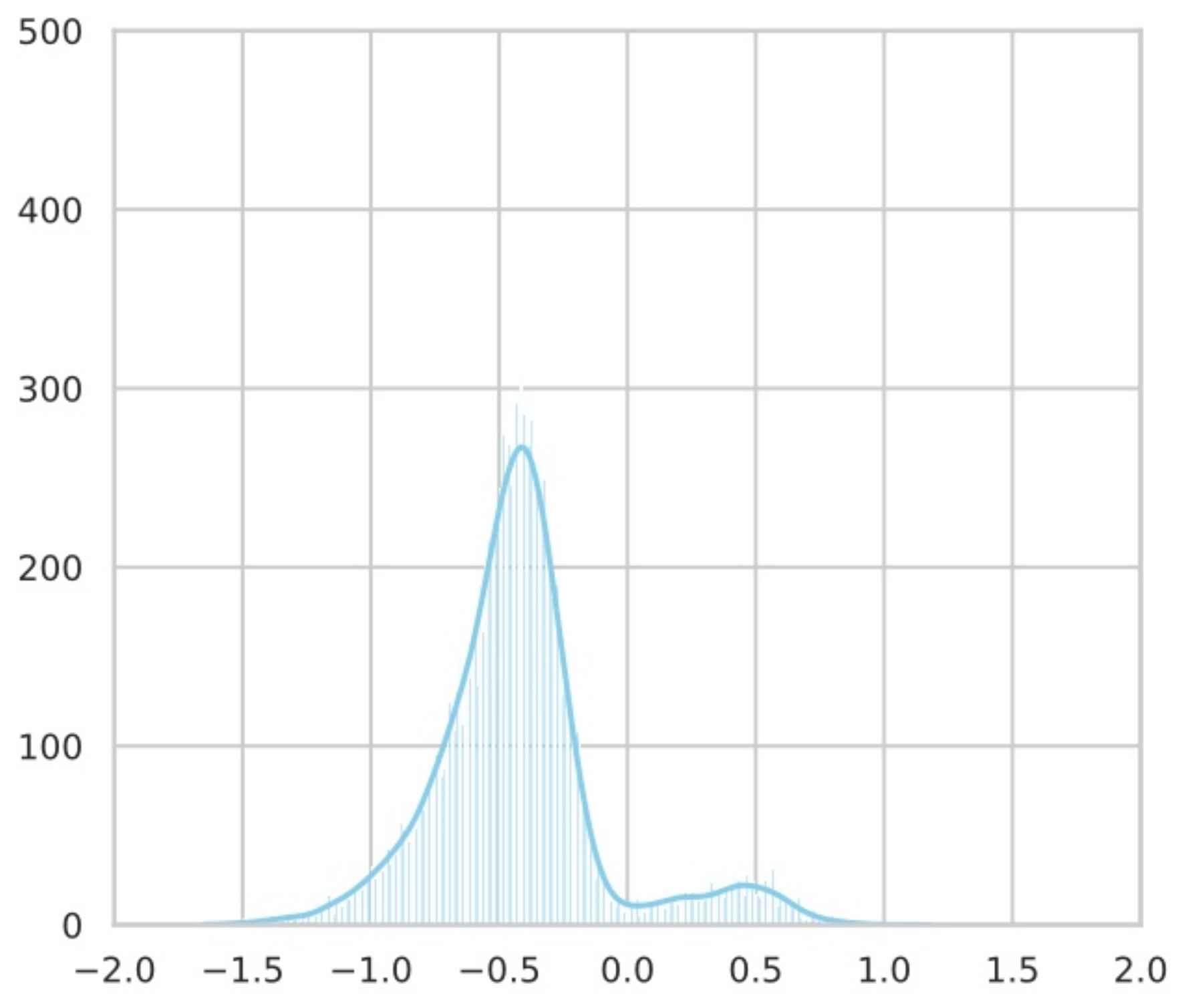}
  \caption{NTD 4,001--5,000}
\end{subfigure}

\vspace{0.6em}

\begin{subfigure}[t]{0.32\textwidth}
  \centering
  \includegraphics[width=\linewidth]{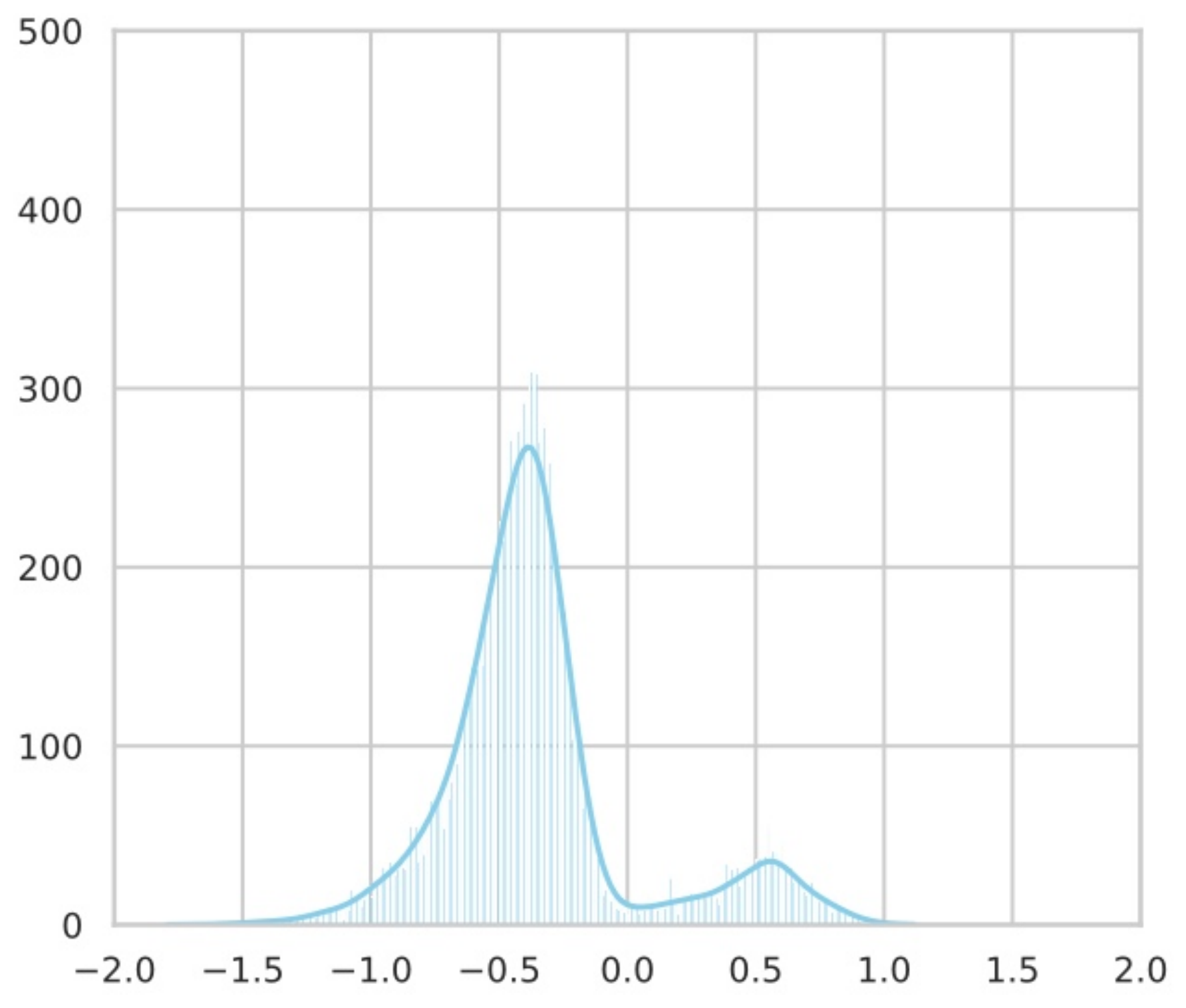}
  \caption{NTD 5,001--10,000}
\end{subfigure}\hfill
\begin{subfigure}[t]{0.32\textwidth}
  \centering
  \includegraphics[width=\linewidth]{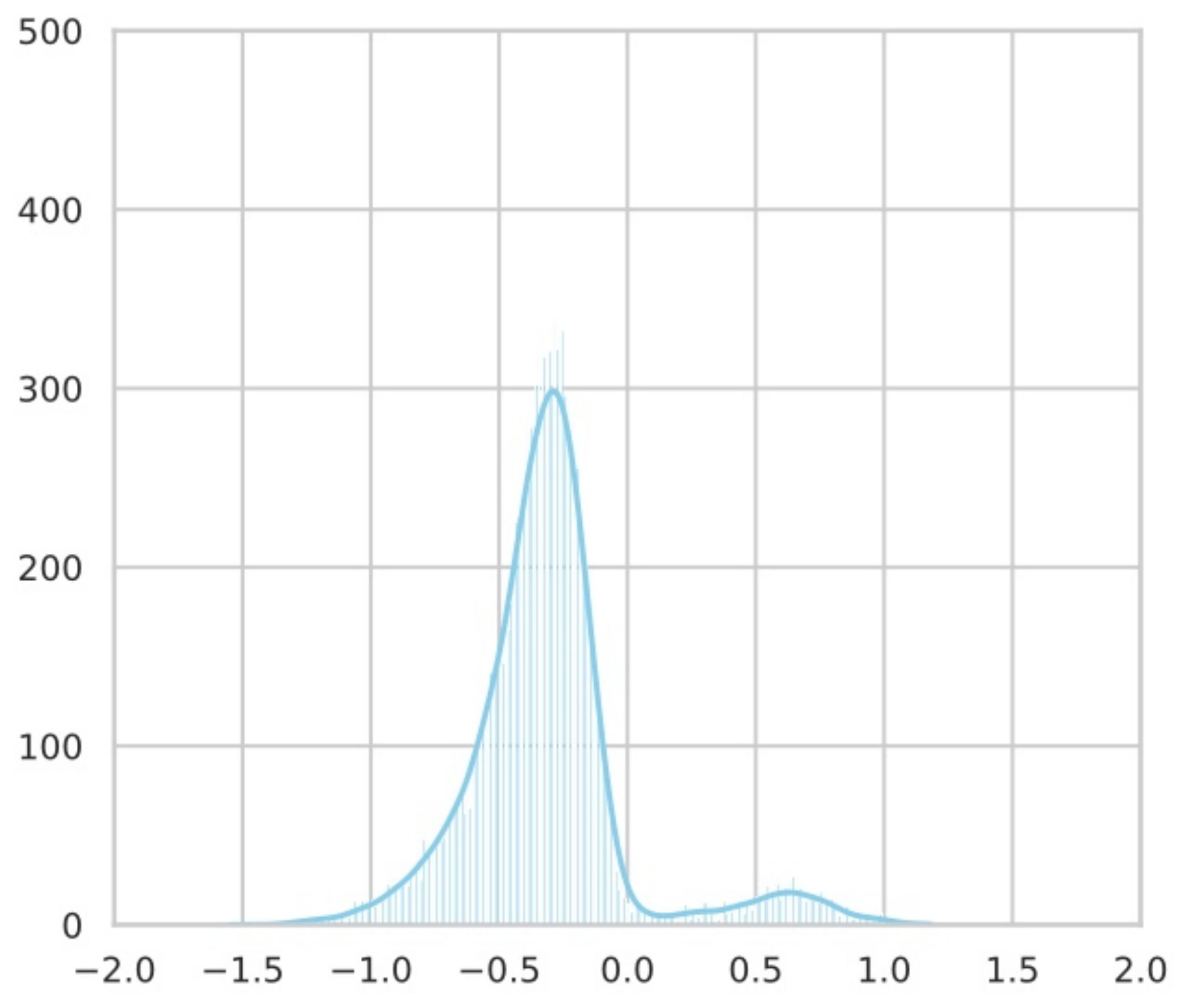}
  \caption{NTD 10,001--20,000}
\end{subfigure}\hfill
\begin{subfigure}[t]{0.32\textwidth}
  \centering
  \includegraphics[width=\linewidth]{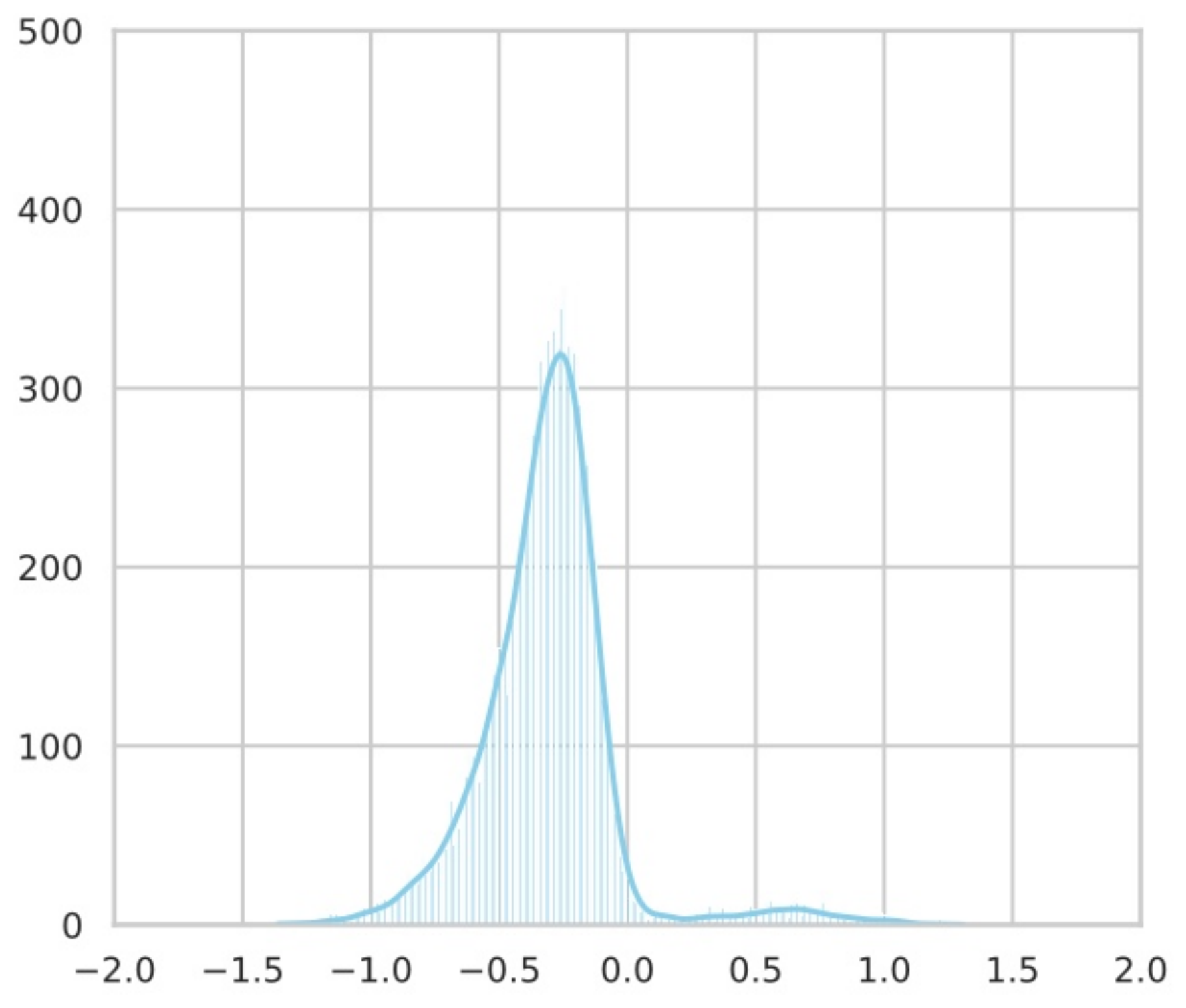}
  \caption{NTD $>$ 20,000}
\end{subfigure}

\caption{Distributions of user-specific price-tier coefficients. Each panel shows the estimated coefficient distribution for one price tier in the ranking model. Using an exchange rate of 1 USD = 32 NTD, these cutoffs correspond to approximately USD 16, 31, 63, 94, 125, 156, 313, 625, and above 625.}
\label{fig:price_coef_dists}
\end{figure*}

We next examine heterogeneity in price-related preferences by turning to the estimated price-tier coefficients \( \pi_p \). Figure~\ref{fig:price_coef_dists} reports the distribution of individual-specific coefficients for each price range. Several features closely parallel the patterns observed for region effects. Most notably, the mass-market tiers (NTD 501--1{,}000 (USD 16--31) and NTD 1{,}001--2{,}000 (USD 31--63)) exhibit pronounced bimodality. As in the case of popular regions such as Bordeaux and Burgundy, these distributions reveal two sizable groups of consumers with opposing preference. In contrast, higher price tiers (above NTD 2{,}001 (USD 63)) display distributions dominated by a negative mode with a small but persistent positive right tail. This mirrors the region results for more niche or specialized origins, where average appeal is limited but a minority segment exhibits strong positive valuation. At the very top end (NTD 10{,}001 and above), the distributions become more concentrated and closer to unimodal, reflecting a narrower effective choice set and stronger regularization arising from sparse interactions. 

\begin{figure}[!htbp]
  \centering
  \includegraphics[width=0.8\textwidth]{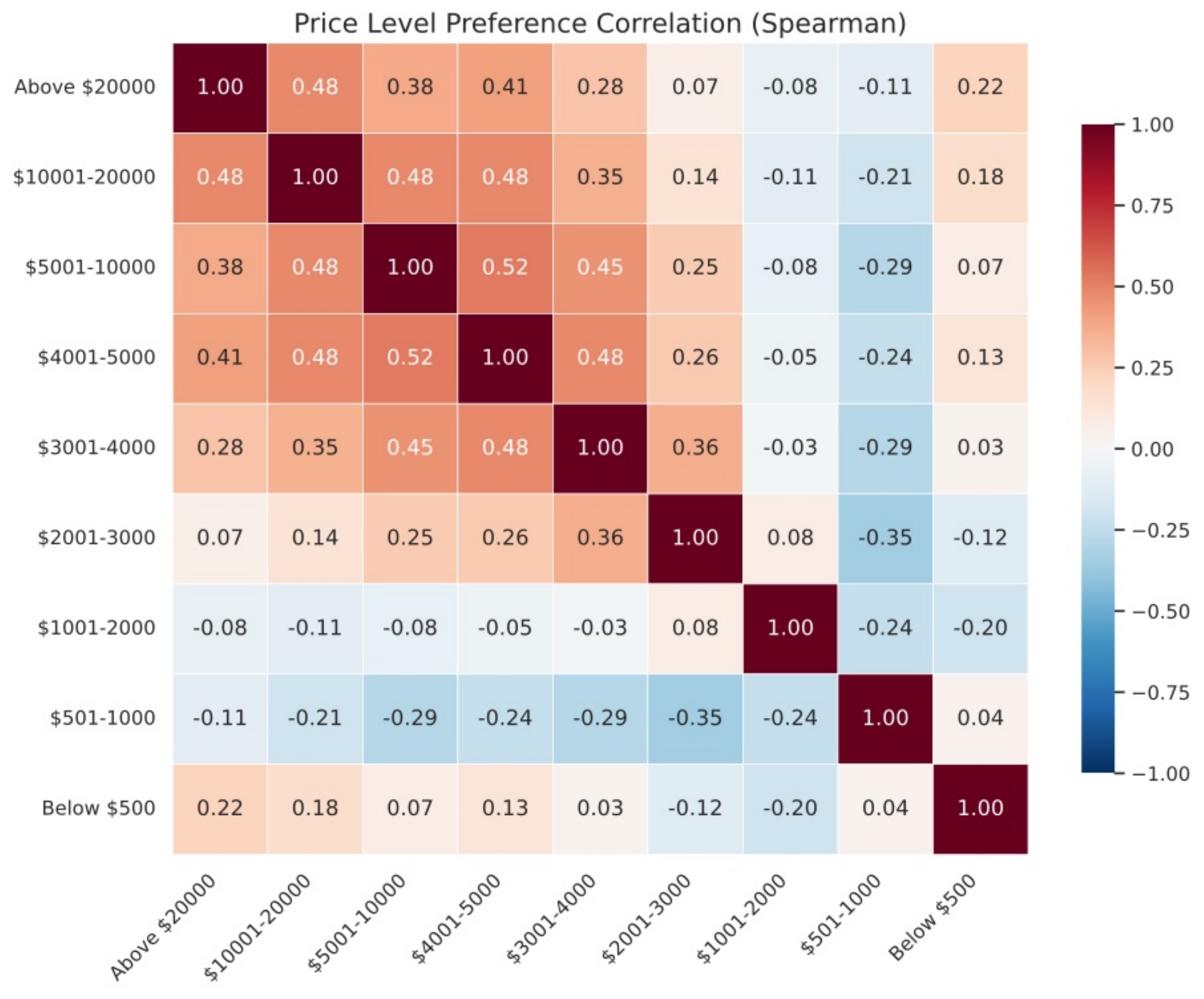}
  \caption{Correlation matrix of price preference effects.
  Each entry reports the pairwise correlation across customers between two price-tier preference effects. Prices shown in the figure are in NTD. Using an exchange rate of 1 USD = 32 NTD, these cutoffs correspond to approximately USD 16, 31, 63, 94, 125, 156, 313, 625, and above 625.}
  \label{fig:price_corr}
\end{figure}

Figure~\ref{fig:price_corr} reports the correlation matrix of the estimated price-tier coefficients. Two broad patterns stand out. First, preferences across higher price tiers exhibit a clear block structure: adjacent premium tiers are strongly positively correlated, with correlations around \(0.48\)–\(0.52\) for neighboring ranges (e.g., NTD~3{,}001--4{,}000 vs.~4{,}001--5{,}000, and 4{,}001--5{,}000 vs.~5{,}001--10{,}000). This indicates that consumers who value premium wines tend to substitute within a narrow band of nearby price tiers rather than focusing on a single exact price point. Correlations decay quickly as price distance increases, suggesting that price-related preferences are organized around local consideration sets rather than a single smooth ranking over the entire price spectrum.

Second, the mass-market tier NTD~501--1{,}000 plays a distinct role. Its coefficients are negatively correlated with most mid- and high-price tiers (e.g., \(\rho=-0.35\) with NTD~2{,}001--3{,}000 and \(\rho=-0.29\) with NTD~3{,}001--4{,}000), indicating that this range acts as a dividing line in consumers’ price preferences. One group systematically favors wines in this range while down-weighting higher tiers, whereas another group does the opposite. By contrast, the very low-price tier (below NTD~500) is weakly correlated with most tiers and even mildly positively correlated with the highest price ranges. This pattern suggests that sub-500 purchases often coexist with premium buying—reflecting add-on, trial, or casual-occasion purchases rather than a strictly low-budget orientation. Overall, the correlation structure reinforces the distributional evidence: price preferences are best described by discrete segmentation of consideration sets, especially around the NTD~501--1{,}000 threshold, with a coherent premium block characterized by substitution among neighboring high-price tiers.

\FloatBarrier

\section{Application I: Recommender System}
\subsection{Recommender System}

We first evaluate the proposed framework in a recommendation setting that focuses explicitly on predicting the purchase of \emph{new} items—that is, wine categories a customer has not previously consumed. This task is of direct managerial relevance: effective recommendations should help customers discover new products rather than merely repeat past purchases. At the same time, predicting new-item adoption is empirically challenging because it requires extrapolating preferences beyond observed consumption histories, especially in markets with highly differentiated products and sparse individual-level purchase data. A key advantage of a ranking-based preference model is precisely its ability to infer relative preferences over unobserved items by combining information from observable attributes and cross-consumer similarity.

Based on the estimated model parameters
\(\hat{\theta} = (\hat{\delta}_r, \hat{\gamma}_g, \hat{\pi}_p, \hat{\lambda}_i, f_{rgp})\),
we construct predicted utility scores for each consumer--item pair,
\[
    \hat{u}_{irgp}
    = \hat{\delta}_r
    + \hat{\gamma}_g
    + \hat{\pi}_p
    + \hat{\lambda}_i^{\top} \hat{f}_{rgp},
\]
which summarize the model-implied relative preference for item \((r,g,p)\) by consumer \(i\). Recommendations are generated by ranking all candidate items according to these predicted scores. For each consumer, we select the top-\(N\) items with the highest predicted utility. To focus on genuine out-of-sample prediction, we exclude from the candidate set any items the consumer has previously purchased. The resulting task therefore evaluates the model’s ability to rank and recommend \emph{previously untried} product categories, rather than its ability to recover observed choices.

Finally, recommendations are constructed at the product-category level rather than at the level of individual stock keeping units. This design choice mirrors the construction of the ranking data used for estimation, where user preferences are modeled over interpretable wine categories defined by region, grape variety, and price tier. Operating at this level allows the model to recommend novel but related products—such as wines from a familiar region or style at a different price point—thereby aligning the recommendation exercise with the underlying preference-learning objective.

\subsection{Evaluation Methods}


To evaluate out-of-sample recommendation performance, we split the transaction data into training and test sets using a time-aware cross-validation procedure. The split is constructed to be strictly disjoint at the transaction level to prevent information leakage. In addition, purchase behavior in wine markets is subject to pronounced short-term cycles arising from seasonality, promotional campaigns, and time-specific product availability. For example, Beaujolais Nouveau is released annually in November and is typically accompanied by intensive, short-lived marketing campaigns that generate sharp and predictable demand spikes.\footnote{Beaujolais Nouveau is traditionally released on the third Thursday of November each year. Retailers often coordinate concentrated promotions and limited-time offerings around the release date, leading to highly localized surges in purchases that are not representative of underlying, stable preferences.} To account for these features of the data, we adopt a splitting strategy that preserves local temporal structure while avoiding systematic distributional differences between the training and test samples.

Specifically, we implement a weekly leave-one-day-out scheme. For each calendar week in the sample period, one weekday is randomly selected and all transactions occurring on that day are assigned to the test set, while transactions from the remaining six days of the same week form the training set. This procedure is applied independently for every week. By construction, the test set draws observations from all weekdays across the sample period, ensuring that evaluation results are not driven by particular days, promotional schedules, or recurring weekly patterns. To make the evaluation exercise feasible, we restrict attention to customers and items that appear in both the training and test samples. If a consumer or product category appears only in one split, out-of-sample prediction is unavailable because the model cannot generate meaningful preference rankings for that unit. Likewise, if a customer appears only in the training data, no out-of-sample evaluation is possible because there are no held-out purchases against which predicted rankings can be assessed.

Recommendation quality is evaluated using Precision@\(K\) and Recall@\(K\), where \(K\) denotes the length of the recommendation list. Both metrics are designed to assess the model’s ability to predict purchases of previously unconsumed product categories. We begin by defining the relevant consumer-specific sets.

For a given consumer \(i\), let \(\mathcal{A}_i\) denote the \emph{ground-truth set} of newly purchased product categories in the test data—that is, categories $j=(r,g,p)$ that appear in the consumer’s test data but were not observed in the training data. Formally,
\[
\mathcal{A}_i
=
\left\{ j=\;\middle|\; j \in \text{Test}_i \;\wedge\; j \notin \text{Train}_i \right\}.
\]
This set captures the consumer’s newly revealed preferences during the evaluation window. Also, let \(\mathcal{B}_i^{(K)}\) denote the \emph{recommendation set}, defined as the top-\(K\) product categories ranked by the model for consumer \(i\), after excluding all categories previously observed in the training data:
\[
\mathcal{B}_i^{(K)} = \{ j_{i1}, j_{i2}, \ldots, j_{iK} \}.
\]
The intersection \(\mathcal{A}_i \cap \mathcal{B}_i^{(K)}\) therefore represents \emph{hits}—categories that are both recommended by the model and subsequently purchased by the consumer in the test data.

Using these definitions, Recall@\(K\) is defined as
\begin{equation}
\text{Recall@}K
=
\frac{\left| \mathcal{A}_i \cap \mathcal{B}_i^{(K)} \right|}
{\left| \mathcal{A}_i \right|},
\end{equation}
which measures the extent to which the recommendation list recovers the consumer’s newly expressed purchase interests. The second evaluation metric Precision@\(K\) is defined as
\begin{equation}
\text{Precision@}K
=
\frac{\left| \mathcal{A}_i \cap \mathcal{B}_i^{(K)} \right|}
{K},
\end{equation}
which is the fraction of recommended items that are subsequently validated by observed purchases. Precision@\(K\) therefore captures ranking accuracy at the top of the recommendation list. Both metrics are computed at the consumer level and then averaged across customers to obtain aggregate performance measures. 

\subsection{Results}
We evaluate the recommendation performance of the proposed model and assess whether combining attribute-based preference components with latent factor learning yields meaningful gains in out-of-sample prediction. All evaluations are conducted at the consumer level using held-out purchase data, following the training and testing protocol described in Section~4.2. Performance is measured using Precision@\(K\) and Recall@\(K\), and statistical significance is assessed via paired tests based on consumer-level metric differences.

Our primary benchmark is a popularity-based recommender that ranks product categories by their overall purchase frequency in the training data. This benchmark reflects a realistic managerial practice in settings where individual-level information is unavailable, incomplete, or too costly to operationalize. In such cases, recommending the most popular items is a simple, low-cost strategy that requires no personalization infrastructure and is therefore easy to adopt in practice. However, because it delivers the same recommendation list to all consumers, this approach ignores heterogeneity in preferences, it may be inefficient when tastes are diverse. Comparing our model against this benchmark allows us to quantify the incremental value of personalization—namely, the extent to which exploiting individual-level preference information can improve recommendation accuracy.

In addition to this benchmark, we consider two restricted variants of our model to isolate the sources of predictive performance. The first variant relies exclusively on observed product attributes—region, grape variety, and price tier—captured by the fixed-effect components \((\delta_r, \gamma_g, \pi_p)\), and therefore exploits only systematic, interpretable differences across products. The second variant relies solely on the latent factor structure \((\lambda_i, f_{rgp})\), abstracting from all observable attributes and capturing preference similarity through collaborative patterns alone. Comparing these restricted specifications to the full model allows us to assess whether predictive gains arise from one component in isolation or from their interaction. As we show below, neither attributes nor latent factors alone are sufficient to match the performance of the full model; rather, it is the combination of structured product information and data-driven latent heterogeneity that delivers the strongest recommendation accuracy.

Table~\ref{tab:rec_performance} reports recommendation performance across different list lengths. First note that popularity-based benchmark in fact performs reasonably well, as the wine market is high-concentrated in Taiwan: a small number of well-known regions and styles—such as Bordeaux and Burgundy—account for a large share of aggregate demand. Nevertheless, the proposed model that combines observed product attributes with latent preference factors consistently outperforms the popularity benchmark. The gains are particularly pronounced for Recall@\(K\) at larger values of \(K\), indicating that personalization is especially valuable when the objective is to recover a broader set of consumers’ newly revealed purchases rather than only the single most likely item. In terms of Precision@\(K\), the proposed model also outperforms the baseline, especially for smaller values of \(K\), indicating more accurate ranking at the top of the recommendation list. This pattern suggests that while popular items serve as a strong baseline, they do not adequately capture the heterogeneity in consumers’ consideration sets. 
\begin{table}[h!]
\centering
\setlength{\tabcolsep}{6pt}
\renewcommand{\arraystretch}{1.4}
\begin{threeparttable}
\caption{Recommendation performance at different list lengths $K$}
\label{tab:rec_performance}
\begin{tabular}{lcccccccc}
\hline \hline
 & \multicolumn{4}{c}{Recall@K} & \multicolumn{4}{c}{Precision@K} \\
\cmidrule(l){2-5}
\cmidrule(l){6-9}
Method & $K{=}1$ & $K{=}10$ & $K{=}20$ & $K{=}40$ 
       & $K{=}1$ & $K{=}10$ & $K{=}20$ & $K{=}40$ \\
\hline
Popularity Benchmark
& 0.0124 & 0.0763 & 0.1241 & 0.1986
& 0.0299 & 0.0245 & 0.0199 & 0.0164 \\ 

Full Model & 0.0125 & 0.0908 & 0.1428 & 0.2140
& 0.0407 & 0.0297 & 0.0242 & 0.0188 \\

Observed Attribute Only 
& 0.0047 & 0.0380 & 0.0667 & 0.1109
& 0.0141 & 0.0127 & 0.0118 & 0.0103 \\

Factor Model Only %
& 0.0038 & 0.0343 & 0.0655 & 0.1110
& 0.0131 & 0.0110 & 0.0104 & 0.0092 \\

\hline \hline
\end{tabular}

\begin{tablenotes}
\footnotesize
\item Notes: The table reports mean Precision@N and Recall@N values averaged across customers.
Recommendations are generated over product categories not previously observed in the training set.
$K$ denotes the length of the recommendation list.
\end{tablenotes}
\end{threeparttable}
\end{table}

While the full model consistently outperforms the benchmark, specifications that rely solely on latent factors or solely on observable attributes deliver similar but weaker performance and are uniformly dominated by the combined specification. This pattern highlights the complementary roles of interaction-based and attribute-based preference information: neither component alone is sufficient to fully exploit the information contained in observed purchase behavior, whereas their combination is essential for capturing the rich structure of heterogeneity in consumer preferences.

\section{Application II: Preference-Based Targeting and Segmentation}
\subsection{From Recommendation to Market-Level Analysis}
Beyond item-level recommendation, the proposed model can also be used to support market-level analysis focused on segmentation and targeting. Rather than evaluating how well the model ranks items for individual users, the goal in this section is to summarize model-implied preference rankings in a way that is informative for aggregate decision-making. By aggregating individual preference information, the model provides a structured representation of how different groups of consumers align with specific product characteristics.

This perspective enables the analysis of questions that fall outside the scope of standard recommender-system evaluations. In particular, it applies to settings in which decisions are made at the segment or product level rather than at the level of individual users. Examples include offline retail environments, where recommendations must be translated into simple rules or guidelines for sales staff, and targeting problems in which firms seek to identify which customer segments are most strongly associated with a given product type or inventory position. In these cases, the analytical focus shifts from ranking products for each consumer to characterizing the relationship between consumer segments and product attributes implied by the estimated preference structure.

\subsection{Target Segment Identification for a Given Product Type}
This subsection addresses the following managerial question: given a specific product type, which customer segments are most likely to exhibit a strong preference for it?

In contrast to conventional recommender-system analyses that focus on ranking items for individual customers, this exercise reverses the perspective by fixing a product type and identifying the customer groups that are disproportionately represented among its most enthusiastic supporters. This formulation is particularly relevant for targeting and inventory-driven decisions, where the managerial problem is “who should this product be marketed to?” rather than “what should be recommended to this customer?”

The analysis is based on customer--product preference scores implied by the estimated recommendation model. These scores are not interpreted in absolute terms. Instead, for each customer, we transform preference scores into within-customer percentile rankings across all product types. This normalization serves two purposes. First, it removes individual-specific scale differences in model scores that arise from heterogeneous interaction intensity or regularization. Second, it yields an ordinal representation of preferences that aligns with the ranking-based nature of the estimation procedure, which identifies relative preference orderings rather than cardinal utility differences. 

Formally, let \(\mathcal{I}\) denote the set of customers and let product types be indexed by \(j = 1,2,\ldots,m\). For each customer \(i \in \mathcal{I}\), let \(s_{ij}\) denote the model-implied preference score for product \(j\). We define the within-customer percentile rank of product \(j\) for customer \(i\) as
\begin{equation}
R_{ij}
=
\frac{\mathrm{rank}\!\left(u_{ij}, \{ u_{ik} \mid k = 1,\ldots,m \}\right)}{m},
\end{equation}
where \(\mathrm{rank}(\cdot)\) assigns lower values to more preferred products. Given a threshold \(\tau \in (0,1)\), we classify customer \(i\) as a \emph{product fan} of product \(j\) if product \(j\) ranks among the top \(\tau\) fraction of that customer’s preferences. The set of product fans for product \(j\) is therefore defined as
\begin{equation}
\mathcal{I}_{j}(\tau)
=
\left\{ i \in \{1,2,...,n\}\;\middle|\; R_{ij} \le  \tau \right\}.
\end{equation}
This set captures customers who exhibit a strong relative preference for product \(j\) according to the model-implied ranking.

The analysis is based on customer--product preference scores implied by the estimated recommendation model. These scores are not interpreted in absolute terms. Instead, for each customer, we transform preference scores into within-customer percentile rankings across all product types. This normalization serves two purposes. First, it removes individual-specific scale differences in model scores that arise from heterogeneous interaction intensity or regularization. Second, it yields an ordinal representation of preferences that aligns with the ranking-based nature of the estimation procedure, which identifies relative preference orderings rather than cardinal utility differences. 

Formally, let \(\mathcal{I}\) denote the set of customers and let product types be indexed by \(j = 1,2,\ldots,m\). For each customer \(i \in \mathcal{I}\), let \(s_{ij}\) denote the model-implied preference score for product \(j\). We define the within-customer percentile rank of product \(j\) for customer \(i\) as
\begin{equation}
R_{ij}
=
\frac{\mathrm{rank}\!\left(u_{ij}, \{ u_{ik} \mid k = 1,\ldots,m \}\right)}{m},
\end{equation}
where \(\mathrm{rank}(\cdot)\) assigns lower values to more preferred products. Given a threshold \(\tau \in (0,1)\), we classify customer \(i\) as a \emph{product fan} of product \(j\) if product \(j\) ranks among the top \(\tau\) fraction of that customer’s preferences. The set of product fans for product \(j\) is therefore defined as
\begin{equation}
\mathcal{I}_{j}(\tau)
=
\left\{ i \in \{1,2,...,n\}\;\middle|\; R_{ij} \le  \tau \right\}.
\end{equation}
This set captures customers who exhibit a strong relative preference for product \(j\) according to the model-implied ranking.

We next quantify how strongly different customer segments are represented among the fans of a given product type using a composition-based lift metric. Let \(S \subseteq \{1,2,...,n\}\) denote a customer segment defined by observable customer attributes, and let \(|S|\) denote the size of the segment. Segments may be defined by a single attribute (e.g., female customers, customers aged 50 and above) or by a combination of attributes (e.g., female customers aged 50 and above with high average spending). 

Given a product type \(j\) and a fan threshold \(\tau\), the \emph{composition lift} of segment \(S\) with respect to product \(j\) is defined as
\begin{equation}
\mathrm{Lift}(S,j)
=
\frac{\Pr(i \in S \mid i \in \mathcal{I}_j(\tau))}{\Pr(i \in S)}
=
\frac{\frac{|S \cap \mathcal{I}_j(\tau)|}{|\mathcal{I}_j(\tau)|}}{\frac{|S|}{n}},
\end{equation}
where $|\cdot|$ is the cardinality of the set (number of customers). This statistic measures whether a given segment is over- or under-represented among customers who exhibit a strong relative preference for product \(j\). A lift greater than one indicates that the segment appears more frequently among product fans than in the overall customer population, while a lift below one indicates under-representation. As a result, it facilitates comparisons across segments of different sizes and avoids mechanically favoring large segments. 

To assess whether observed lift values reflect systematic concentration rather than sampling variation, we conduct a binomial test for each segment--product pair. The null hypothesis assumes that customers who are fans of product \(j\) are drawn randomly from the overall population, with success probability equal to the segment’s population share \(|S|/n\). This procedure helps distinguish economically meaningful over-representation from noise, especially when evaluating smaller or more finely defined segments. 

\begin{figure}[htbp]
    \centering
    
    \begin{subfigure}[b]{\textwidth}
        \centering
        \includegraphics[width=\linewidth]{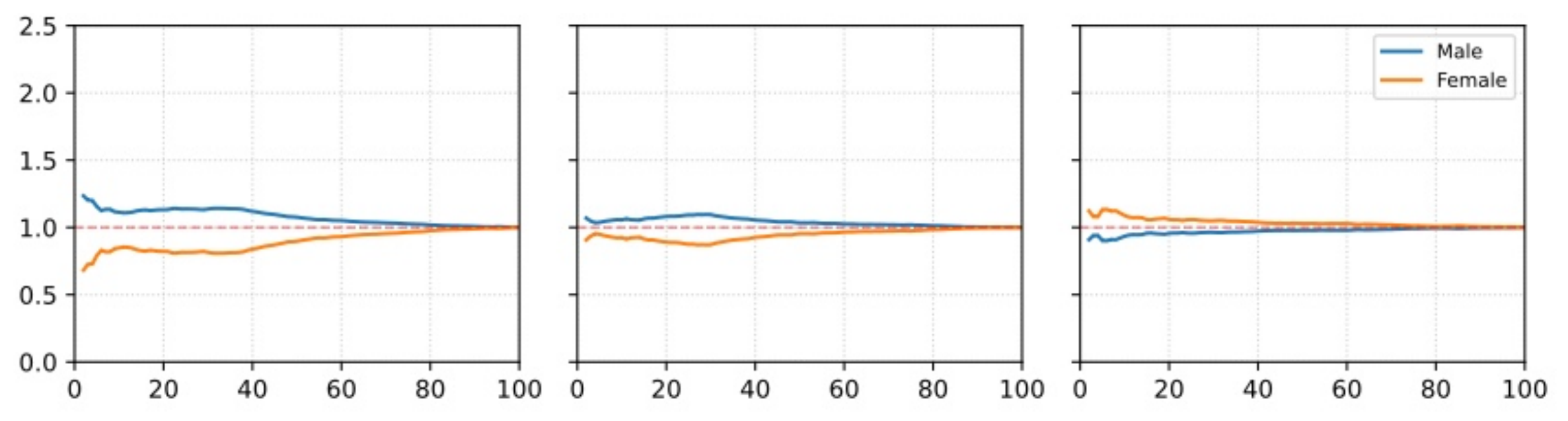}
        \caption{Segmentation by Gender}
    \end{subfigure}
    
    \vspace{0.5cm}
    
    \begin{subfigure}[b]{\textwidth}
        \centering
        \includegraphics[width=\linewidth]{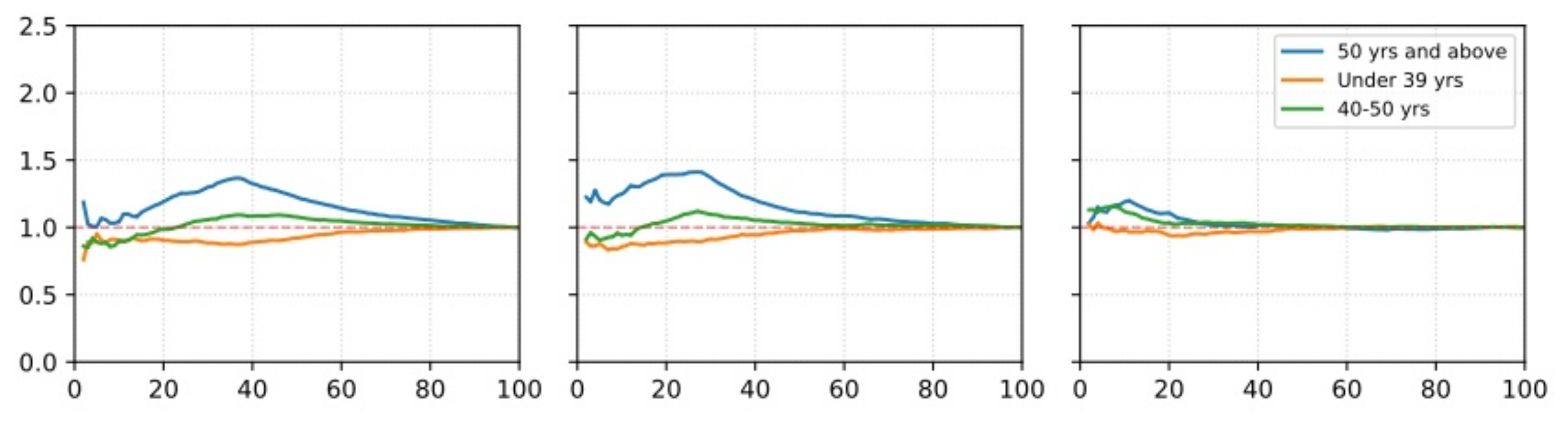}
        \caption{Segmentation by Age Group}
    \end{subfigure}
    
    \vspace{0.5cm}
    
    \begin{subfigure}[b]{\textwidth}
        \centering
        \includegraphics[width=\linewidth]{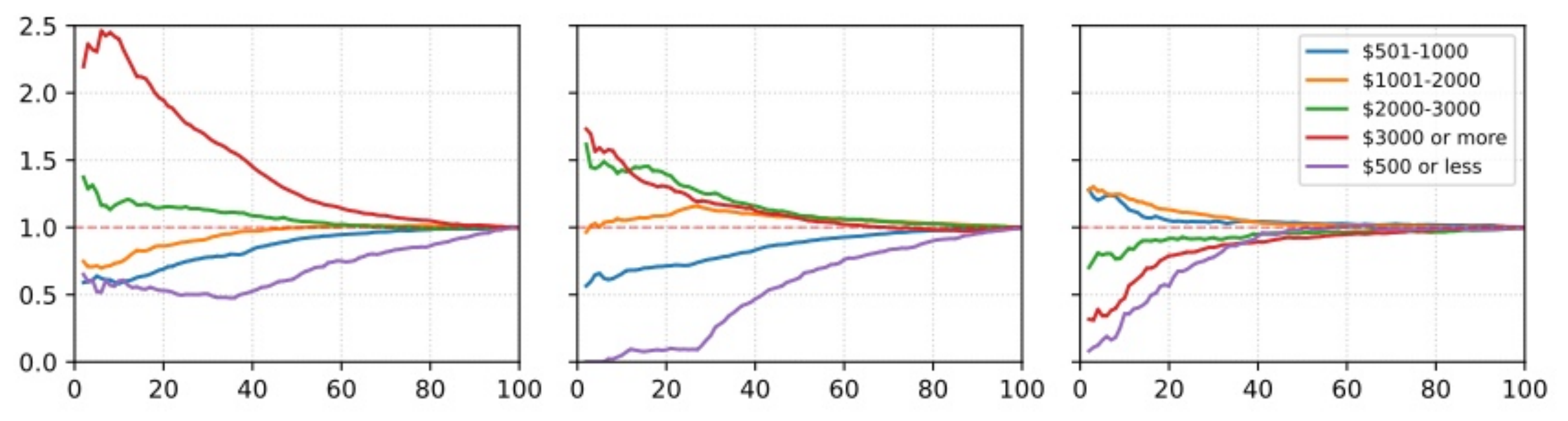}
        \caption{Segmentation by Average Price per Bottle (APB)}
    \end{subfigure}
    
    \caption{Lift curve analysis across three key wine regions: Bordeaux (Left column), Burgundy (Middle column), and Marlborough (Right column), segmented by (a) Gender, (b) Age Group, and (c) Average Price per Bottle (APB). The x-axis represents the threshold $q$, defining the set of \emph{product fans} as the top $q\%$ of customers who rank the region highest within their personal preferences. The y-axis displays the composition lift for each segment at that threshold, quantifying the segment's over-representation among fans relative to its share in the general population. The red dashed line ($y=1$) indicates the baseline where a segment's representation among fans equals its population share.}
    
    \label{fig:lift_apb}
\end{figure}

Figure~\ref{fig:lift_apb} reports the composition lift analysis by examining how preference concentration evolves from the most extreme top-ranked customers to broader segments. Rather than fixing a single cutoff, this approach traces lift as a function of the fan threshold $\tau$ providing a distributional view of preference concentration across customer segments.

Two general patterns emerge from the empirical results. First, consistent with the methodology's premise, Figure~\ref{fig:lift_apb}(c) demonstrates that for premium products, lift is strongly elevated among small values of $q$. This is most evident in the Old World regions (Bordeaux and Burgundy), where the ``\$3000 or more'' spending group (red line) exhibits a lift exceeding 2.0 at the strictest fan thresholds ($q < 10\%$). As $q$ increases, the lift curve converges toward one. This pattern confirms that the appeal of these regions is driven by a relatively small but highly committed audience of high-spending collectors. Conversely, Marlborough (Figure~\ref{fig:lift_apb}(c), Right) displays an inverted pattern where lower-to-mid price tiers (\$501--2000) show the highest lift, accurately reflecting its market positioning as an accessible, value-driven region.

\begin{figure}[!htbp]
    \centering
    
    \begin{subfigure}[b]{\textwidth}
        \centering
        \includegraphics[width=\linewidth]{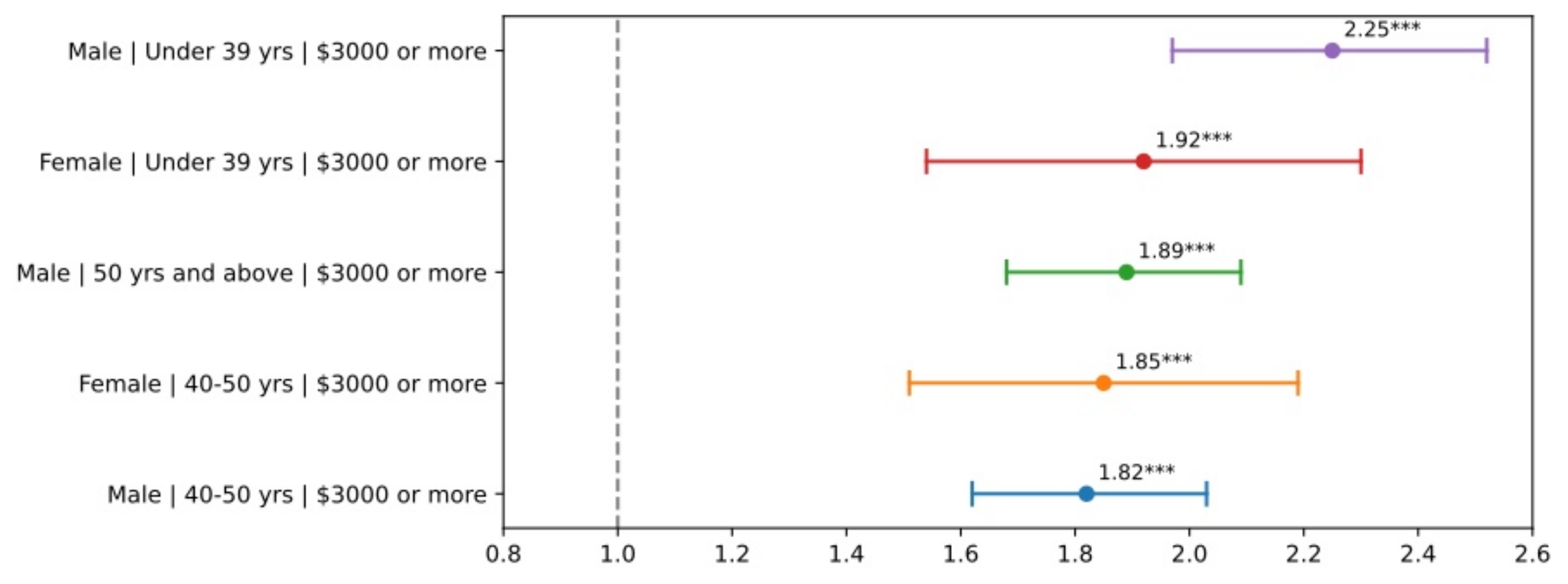}
        \caption{Top Target Segments for Bordeaux}
    \end{subfigure}
    
    \vspace{0.5cm}
    
    \begin{subfigure}[b]{\textwidth}
        \centering
        \includegraphics[width=\linewidth]{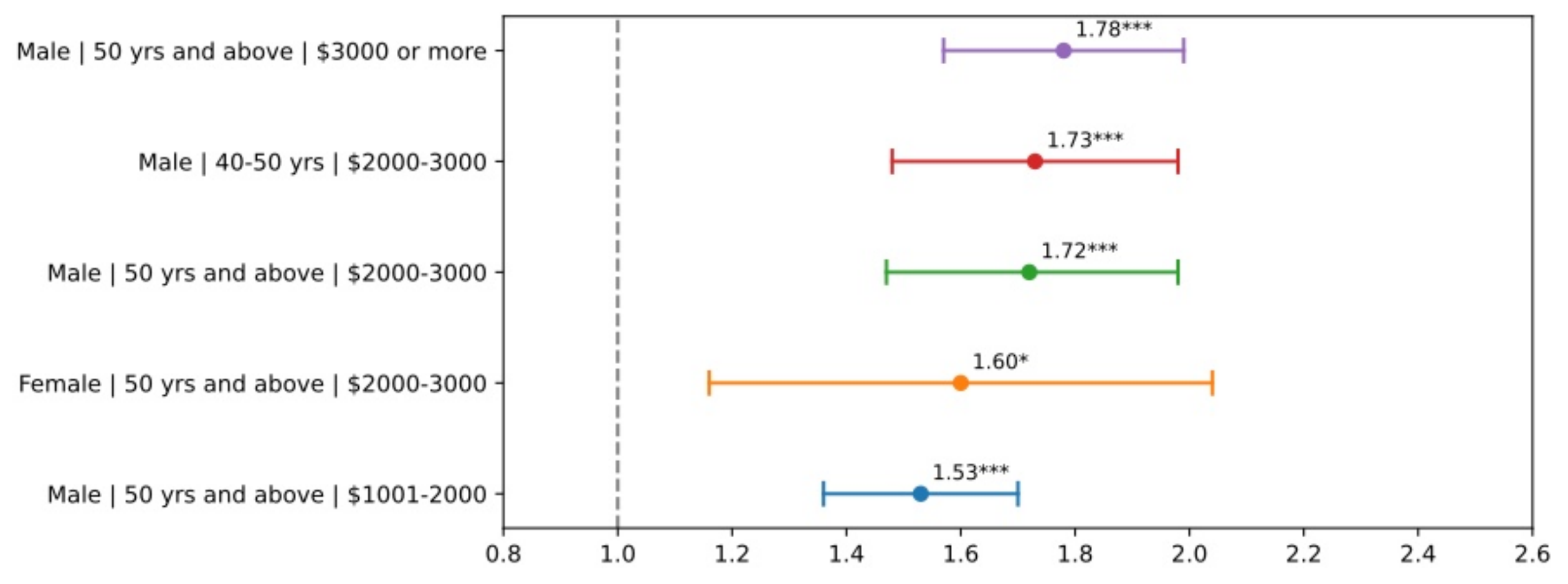}
        \caption{Top Target Segments for Burgundy}
    \end{subfigure}
    
    \vspace{0.5cm}
    
    \begin{subfigure}[b]{\textwidth}
        \centering
        \includegraphics[width=\linewidth]{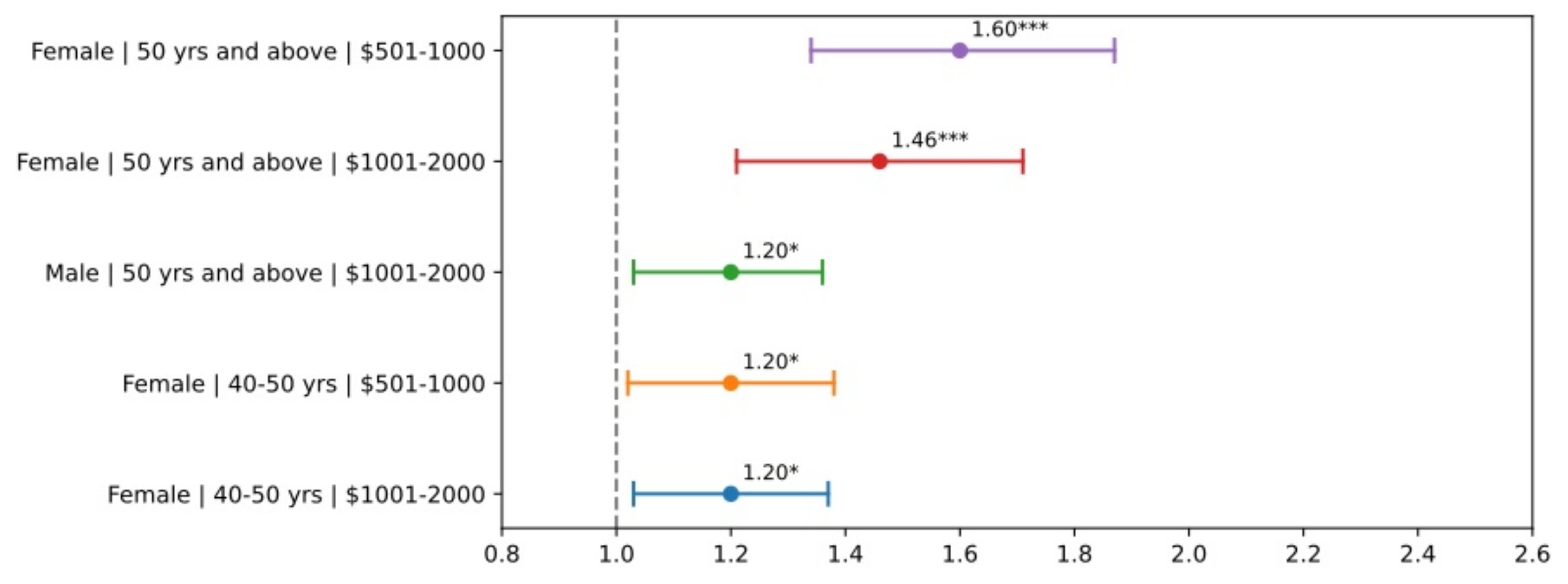}
        \caption{Top Target Segments for Marlborough}
    \end{subfigure}
    
    \caption{Identification of the top five customer segments exhibiting the strongest preference for each region. The x-axis represents the composition lift, quantifying the segment's over-representation among product fans. The point estimates of the lift scores are accompanied by horizontal bars representing the 95\% confidence intervals from the binomial test, indicating the statistical reliability of the preference signal.}
    \label{fig:target_segments_analysis}
\end{figure}

Second, the analysis of specific segment intersections in Figure~\ref{fig:target_segments_analysis} reveals non-monotonic preference structures that conventional mean-based models would likely obscure. An interesting example is observed in the Bordeaux analysis. In the aggregate univariate analysis (Figure~\ref{fig:lift_apb}(b), Left), the ``Under 39 yrs'' age group (orange line) generally underperforms, appearing below the baseline ($y=1$). A standard regression model might thus conclude that younger customers possess a weak affinity for Bordeaux. However, the multivariate ranking analysis in Figure~\ref{fig:target_segments_analysis}(a) reveals that the specific intersection of {Male $|$ Under 39 yrs $|$ \$3000 or more} is actually the \textit{highest-performing} segment (Lift = 2.25). This reversal highlights the model's ability to identify ``niche but strategically valuable groups''---in this case, young, affluent collectors---who are statistically invisible when attributes are averaged independently.

Furthermore, the comparison between Figure~\ref{fig:target_segments_analysis}(b) (Burgundy) and Figure~\ref{fig:target_segments_analysis}(c) (Marlborough) highlights the model's capacity to handle heterogeneous preference structures without manual product engineering. The top segments for Burgundy are strictly dominated by {Male $|$ 50 yrs+ $|$ \$3000+} customers (Lift = 1.78), reflecting a traditionalist, high-capital profile. In stark contrast, Marlborough's top segments are defined by {Female $|$ 50 yrs+ $|$ \$501--1000} customers (Lift = 1.60). This distinction proves that preferences are not uniformly ordered along a single intensity dimension; rather, the model successfully disentangles the ``Exclusive/Premium'' structure of French wines from the ``Daily/Value'' structure of New World wines.


\clearpage
\section{Conclusion}

This paper develops a flexible framework for learning individual preferences from incomplete ranking data. By interpreting observed rankings as collections of pairwise comparisons, the proposed approach combines interpretable product attributes with a latent factor structure to recover heterogeneous preference orderings across consumers. The model is designed for settings in which complete rankings are unavailable and observed choices provide only partial information about underlying preferences.

Using transaction data from an online wine retailer, we illustrate how the framework can be applied to infer preferences over product categories defined by region, grape variety, and price tier. The empirical results document substantial heterogeneity in both origin and price preferences, including bimodal patterns and discrete segmentation in consumers’ consideration sets. In out-of-sample recommendation tasks, the model consistently outperforms a popularity-based benchmark and restricted specifications that rely solely on observable attributes or latent factors, highlighting the value of combining structured product information with data-driven preference heterogeneity.

Beyond recommendation, we show how model-implied preference rankings can be aggregated to support market-level analysis. By defining product fans and constructing composition-based lift measures, the framework enables firms to identify which customer segments are most strongly aligned with specific product types and to characterize segment-level preference profiles. These applications address managerial questions that standard recommender systems are not designed to answer, particularly in targeting, segmentation, and offline decision contexts.



\bibliography{reference}
\newpage
\begin{appendices}
\section{Wine Flavor Style Categorization}
To represent wine styles in a way that is both comparable across regions and suitable for the following analysis, this study constructs a unified wine style taxonomy using a rule-based procedure. The taxonomy is designed to address two data limitations that are particularly consequential in empirical marketing settings. First, grape variety nomenclature is not standardized across producers and origins: synonyms, alternative spellings, and region-specific aliases frequently refer to the same cultivar. If taken at face value, such discrepancies would mechanically inflate product differentiation and contaminate cross-region comparisons of consumer demand with measurement noise. Second, the space of variety compositions exhibits extreme granularity: many grape sets appear only a few times in the catalog, so treating each distinct composition as its own style would generate a high-cardinality categorical structure, induce substantial sparsity, and yield imprecise inference due to fragmented support. By standardizing variety names, mapping recognized compositions to canonical global styles, and consolidating low-frequency residual categories within each region into broader color-based groups, the proposed taxonomy preserves interpretable dimensions of horizontal product differentiation while reducing spurious heterogeneity and improving the statistical precision of downstream estimates.

In consultation with iCheers and with reference to their cataloging conventions, we then implement the following rule-based classification procedure to address these limitations.

\paragraph{Step 1. Standardize grape variety names.}
We first clean the raw grape variety field by mapping synonyms and region-specific aliases to a
single standardized variety name (e.g., alternative spellings or local names referring to the same
cultivar are merged). After this step, each wine is represented by a standardized set of grape
varieties. In particular, alternative spellings such as ``Moscato'' and ``Muscat'' are both mapped to \emph{Muscat}; ``Pinot Gris'' and ``Pinot Grigio'' are mapped to a single standardized name (e.g., \emph{Pinot Grigio}), so that the two labels are treated as the same grape variety.

\paragraph{Step 2. Create an initial, variety-set style label.}
For each wine, we extract a preliminary set based only on the \emph{set} of
standardized grape varieties it contains, ignoring blending proportions. For example, a wine
made from $\{Cabernet\ Sauvignon, Merlot\}$ receives the same preliminary label regardless of
the relative shares of the two grapes. Single-varietal wines are represented by a singleton set
(e.g., $\{Chardonnay\}$). For instance, regardless of blending proportions, both a 70/30 blend and a 50/50 blend of Cabernet Sauvignon and Merlot both receive the same label $\{Cabernet\ Sauvignon, Merlot\}$; a 100\% Chardonnay wine receives $\{Chardonnay\}$.

\paragraph{Step 3. Map to canonical global styles (if applicable).}
We then apply a predefined rulebook that links certain grape compositions to canonical global wine
styles (e.g., well-known blends or varietal styles). The set of canonical styles and their corresponding
rules is constructed through iterative discussions with the industry partner, reflecting both domain
expert knowledge and practical relevance in real-world wine categorization. Specifically, for each
wine, we evaluate whether its standardized grape set satisfies the criteria of any canonical style
(See Table 5 for the full rule list):
\begin{itemize}
  \item If a match is found, we assign the corresponding canonical style label.
  \item If no match is found, we keep the wine's preliminary variety-set label from Step 2 for single-varietal wine. For multi-varietal wine, they are treated as minor and are collapsed into a broader category — \textquotedblleft red wine\textquotedblright\ or
  \textquotedblleft white wine\textquotedblright\ — based on wine color.
\end{itemize}

The decision to collapse non-canonical multi-varietal blends into broad color-based categories is
motivated by data sparsity considerations. Such blends are typically less standardized and exhibit
highly fragmented purchase frequencies, which can substantially increase feature dimensionality
and impair the robustness of downstream preference estimation. By aggregating these rare and
idiosyncratic blends into coarse categories, we reduce dimensionality while preserving meaningful
signals related to general wine preferences.

For instance, if the rulebook recognizes $\{\text{Grenache}, \text{Syrah}, \text{Mourv\`edre}\}$ as the canonical ``GSM''
style, then any wine with that standardized set is labeled ``GSM''; if $\{\text{Carm\`en\`ere}\}$ has no
canonical match, a single-varietal Carm\`en\`ere wine retains the label $\{\text{Carm\`en\`ere}\}$; if
$\{Tempranillo, Graciano\}$ has no canonical match, the wine is collapsed to
\textquotedblleft red wine\textquotedblright\ (or to \textquotedblleft white wine\textquotedblright\
for an unmatched white blend).


\paragraph{Step 4. Collapse rare styles within each region.}
To avoid an excessively sparse set of style categories, we perform a region-specific consolidation.
Within each region, we count the number of distinct products associated with each style label
obtained from Step 3 and rank styles by these counts. We define \emph{major styles} as those with
more than 10 distinct products, and we retain the most common styles until their cumulative product
share reaches 80\% within that region:
\begin{itemize}
  \item Wines belonging to major styles retain their labels.
  \item All remaining styles are treated as \emph{minor} and are collapsed into a broader category
  --- \textquotedblleft red wine\textquotedblright\ or \textquotedblleft white wine\textquotedblright\ ---
  based on wine color.
\end{itemize}

Within a given region, suppose \textquotedblleft GSM\textquotedblright\ and $\{Syrah\}$ each have more than 10 distinct products and together account for at least 80\% of products; these labels are retained as major styles, while all other low-frequency labels in that region (e.g., rare single-varietal or residual labels) are pooled into \textquotedblleft red wine\textquotedblright\ or \textquotedblleft white wine\textquotedblright\ according to color. As a result, the final taxonomy keeps frequent, interpretable style labels while pooling rare
categories to improve statistical reliability in downstream modeling.
Table~4 provides several examples of the mapping from raw grape-composition inputs to the final variety-group classifications used in our analysis.

\begin{table}[!htbp]
\centering
\caption{Original Grape Composition and Final Variety-Group Classification}
\label{tab:grape_ratio_final_class_en}
\begin{tabularx}{\linewidth}{>{\raggedright\arraybackslash}X >{\raggedright\arraybackslash}p{4.0cm}}
\toprule
Original Grape Composition & Final Variety-Group Classification \\
\midrule
Chardonnay 35\%, Pinot Meunier 5\%, Pinot Noir 65\% & Champagne Blend \\
Pinot Noir 100\% & Pinot Noir \\
Riesling 60\%, S\'emillon 40\% & White wine \\
Chardonnay 100\% & White wine \\
Cinsault / Grenache (proportions unknown) & Ros\'e wine \\
Pinot Noir / Poulsard (Ploussard) / Trousseau (proportions unknown) & Red wine \\
Muscat 100\% & Muscat Sparkling \\
Vermentino 100\% & White wine \\
Grenache 70\%, Syrah 8\%, Mourv\`edre 8\%, Carignan 8\%, Other 6\% & GSM Blend \\
Macabeo 50\%, Xarel-lo 25\%, Parellada 20\%, Mourv\`edre 5\% & Cava Blend \\
\bottomrule
\end{tabularx}
\end{table}


\begin{table}[htbp]
\centering
\setlength{\tabcolsep}{5pt}
\renewcommand{\arraystretch}{1.3}
\caption{Rule-based classification scheme for wine style identification}
\label{tab:wine_rules}
\begin{threeparttable}
\begin{tabular}{p{3.2cm} p{4.2cm} p{3.8cm} p{3.2cm}}
\hline \hline
\textbf{Wine Style} & \textbf{Core Varieties} & \textbf{Classification Rule} & \textbf{Notes} \\
\hline

Bordeaux Blend White &
Sauvignon Blanc, Sémillon, Muscadelle &
At least two core varieties; direct match if labeled Bordeaux Blend and white &
White Bordeaux-style blend \\

Pinot Noir Sparkling &
Pinot Noir &
Must include Pinot Noir; sparkling category &
Single-variety sparkling \\

Chardonnay Sparkling &
Chardonnay &
Must include Chardonnay; sparkling category &
Single-variety sparkling \\

Muscat d'Asti &
Moscato Bianco &
Must include Moscato Bianco &
Sweet sparkling style \\

Prosecco &
Glera &
Must include Glera &
Prosecco definition \\

Cabernet Sauvignon--Syrah Blend &
Cabernet Sauvignon, Syrah &
At least two core varieties &
Syrah and Shiraz treated as equivalent \\

Grenache--Tempranillo Blend &
Grenache, Tempranillo &
At least two core varieties &
No proportion restriction \\

Petite Sirah--Zinfandel Blend &
Petite Sirah, Zinfandel &
At least two core varieties &
Petite Sirah not interchangeable with Syrah \\

Marsanne--Roussanne Blend &
Marsanne, Roussanne &
At least two core varieties &
Classic Rhône white blend \\

Bordeaux Blend Red &
Cabernet Sauvignon, Merlot, Cabernet Franc, Petit Verdot, Malbec, Carménère &
At least two core varieties &
Global Bordeaux-style red blend \\

Champagne Blend &
Chardonnay, Pinot Noir, Pinot Meunier &
At least two core varieties; sparkling category &
Champagne-style blend \\

Rhône Blend &
Grenache, Syrah, Mourvèdre, Carignan, Cinsault &
At least two core varieties &
GSM as representative subset \\

GSM Blend &
Grenache, Syrah, Mourvèdre &
Must include all three varieties &
Canonical GSM definition \\

Super Tuscan Blend &
Sangiovese + international varieties &
Must include Sangiovese and at least one international variety &
No fixed blending ratio \\

Port Blend &
Touriga Nacional, Touriga Franca, Tinta Roriz, Tinta Barroca, Tinto C\~ao &
At least two core varieties; fortified wine &
Douro Valley tradition \\

Chianti Blend &
Sangiovese, Canaiolo, Colorino &
At least two core varieties &
Tuscan red blend \\

Cava Blend &
Macabeo, Parellada, Xarel-lo &
At least two core varieties; sparkling category &
Traditional Cava blend \\

Amarone Blend &
Corvina Veronese, Corvinone, Rondinella &
At least two core varieties &
Amarone-style composition \\

Valpolicella Blend &
Corvina Veronese, Molinara, Rondinella &
At least two core varieties &
Valpolicella-style blend \\

Passetoutgrain Blend &
Gamay, Pinot Noir &
At least two core varieties &
Traditional Burgundy blend \\

\hline \hline
\end{tabular}
\end{threeparttable} 
\end{table}
\end{appendices}

\end{document}